\documentclass{article}

\usepackage{microtype}
\usepackage{graphicx}
\usepackage{subcaption}
\usepackage{booktabs} % for professional tablesCBR-
\usepackage{newunicodechar}
\usepackage{multirow}
\usepackage{booktabs} 
\usepackage{colortbl} 
\usepackage{xcolor}
\usepackage[table]{xcolor}
\usepackage{colortbl}
\usepackage{pgf}
\usepackage{tcolorbox}
\usepackage{stfloats}

\newunicodechar{−}{\textminus}
\usepackage{hyperref}

\NewDocumentCommand{\heng}
{ mO{} }{\textcolor{red}{\textsuperscript{\textit{Heng}}\textsf{\textbf{\small[#1]}}}}

\definecolor{bluep}{rgb}{0.2, 0.2, 0.6}
\NewDocumentCommand{\pj}
{ mO{} }{\textcolor{bluep}{\textsuperscript{\textit{PatrickJ}}\textsf{\textbf{\small[#1]}}}}

\usepackage[preprint]{icml2026}

\usepackage{amsmath}
\usepackage{amssymb}
\usepackage{mathtools}
\usepackage{amsthm}

\definecolor{morandiblue}{RGB}{200,210,225}
\definecolor{morandired}{RGB}{232,205,200}
\definecolor{groupgray}{gray}{0.95}

\newcommand{\shadeperf}[2]{%
  \begingroup
  \pgfmathsetmacro{\absd}{abs(#2)}%
  \pgfmathtruncatemacro{\pctint}{min(99, max(8, round(15 + 4*\absd)))}%
  \ifdim #2pt > 0pt
    \edef\temp{\noexpand\cellcolor{morandiblue!\pctint}}%
    \temp#1%
  \else\ifdim #2pt < 0pt
    \edef\temp{\noexpand\cellcolor{morandired!\pctint}}%
    \temp#1%
  \else
    #1%
  \fi\fi
  \endgroup
}

\usepackage[capitalize,noabbrev]{cleveref}

\theoremstyle{plain}

\theoremstyle{definition}

\theoremstyle{remark}

\usepackage[textsize=tiny]{todonotes}

\DeclareUnicodeCharacter{266B}{\textmusicalnote}
\begin{document}

\twocolumn[
  \icmltitle{Rethinking the Reranker: Boundary-Aware Evidence Selection for Robust Retrieval-Augmented Generation}
%\texttt{BAR-RAG}
%\texttt{BAR-RAG}
    % Rethinking the Reranker: Boundary-Aware Evidence Selection for Robust Retrieval-Augmented Generation



  \icmlsetsymbol{equal}{*}
  \begin{icmlauthorlist}
    \icmlauthor{Jiashuo Sun}{uiuc}
    \icmlauthor{Pengcheng Jiang}{uiuc}
    \icmlauthor{Saizhuo Wang}{hkust}
    \icmlauthor{Jiajun Fan}{uiuc}
    \icmlauthor{Heng Wang}{uiuc}
    \icmlauthor{Siru Ouyang}{uiuc}
    \icmlauthor{Ming Zhong}{uiuc}
    \icmlauthor{Yizhu Jiao}{uiuc}
    \icmlauthor{Chengsong Huang}{washu}
    \icmlauthor{Xueqiang Xu}{uiuc}
    \icmlauthor{Pengrui Han}{uiuc}
    \icmlauthor{Peiran Li}{tamu}
    \icmlauthor{Jiaxin Huang}{washu}
    \icmlauthor{Ge Liu}{uiuc}
    \icmlauthor{Heng Ji}{uiuc}
    \icmlauthor{Jiawei Han}{uiuc}
  \end{icmlauthorlist}

  \icmlaffiliation{uiuc}{University of Illinois Urbana-Champaign}
  \icmlaffiliation{hkust}{Hong Kong University of Science and Technology}
  \icmlaffiliation{washu}{Washington University in St. Louis}
  \icmlaffiliation{tamu}{Texas A\&M University}

  \icmlcorrespondingauthor{Jiashuo Sun}{jiashuo5@illinois.edu}

  \icmlkeywords{Machine Learning, ICML}

  \vskip 0.3in
]



\printAffiliationsAndNotice{}  % no special notice (required even if empty)

\begin{abstract}
Retrieval-Augmented Generation (RAG) systems remain brittle under realistic retrieval noise, even when the required evidence appears in the top-$K$ results.
A key reason is that retrievers and rerankers optimize solely for relevance, often selecting either trivial, answer-revealing passages or evidence that lacks the critical information required to answer the question, without considering whether the evidence is suitable for the generator.
We propose \texttt{BAR-RAG}, which reframes the reranker as a boundary-aware evidence selector that targets the generator’s Goldilocks Zone—evidence that is neither trivially easy nor fundamentally unanswerable for the generator, but is challenging yet sufficient for inference and thus provides the strongest learning signal.
\texttt{BAR-RAG} trains the selector with reinforcement learning using generator feedback, and adopts a two-stage pipeline that fine-tunes the generator under the induced evidence distribution to mitigate the distribution mismatch between training and inference.
Experiments on knowledge-intensive question answering benchmarks show that \texttt{BAR-RAG} consistently improves end-to-end performance under noisy retrieval, achieving an average gain of 10.3\% over strong RAG and reranking baselines while substantially improving robustness. The code is avaliable at \url{https://github.com/GasolSun36/BAR-RAG}.
\end{abstract}

\begin{figure}[t]
    \centering
    \includegraphics[width=0.85\linewidth]{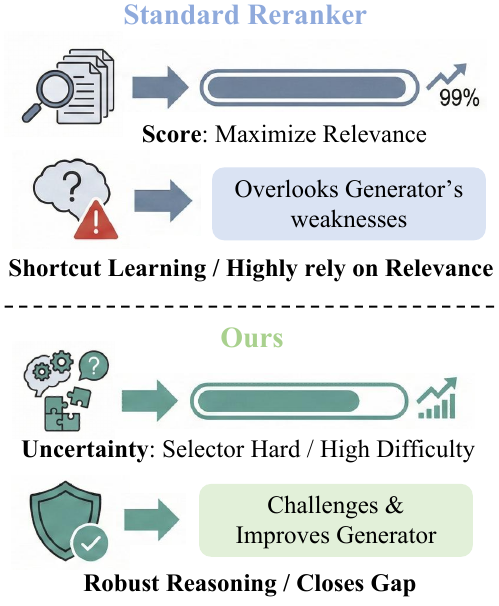}
    \caption{Comparison between standard relevance-based rerankers and our boundary-aware evidence selection.
Standard rerankers maximize relevance scores but overlook the generator’s weaknesses, often encouraging shortcut learning and brittle reasoning by prioritizing trivial or answer-revealing evidence.
In contrast, our method selects challenging yet solvable evidence based on generator uncertainty, promoting robust reasoning and reducing the mismatch between the evidence distributions encountered during training and inference under noisy retrieval.}
    \label{fig:intro}
\end{figure}

% \jh{For drawing, it could be better to say: upper one: "Shortcut Learning/Highly rely on Relevant" →\rightarrow "Learning on easy  cases leads to weak re-ranker";
% bottom: "→\rightarrow  Training on hard/boundary cases leads to robust reasoning"
% }
% \sjs{ Seems we just want to illustrate our Paradigm is better, so avoid use "weak" reranker.}

\section{Introduction}

Retrieval-Augmented Generation (RAG) has achieved remarkable success on knowledge-intensive tasks by grounding large language model (LLM) outputs in retrieved evidence \cite{rag_survey, agentic_ai}. Yet RAG systems remain surprisingly brittle when retrieval results are noisy, partially relevant, or requires multi-step integration, even when the necessary facts exist somewhere in the top-$K$ results \cite{ragged, rankrag, power_of_noise}. In such realistic settings, LLMs often fail to synthesize scattered information and instead hallucinate plausible but incorrect answers. This fragility reveals a fundamental limitation: retrievers are optimized for relevance, not for providing evidence that maximally strengthens the generator's reasoning.

\begin{figure*}[t]
    \centering
    \includegraphics[width=0.48\textwidth]{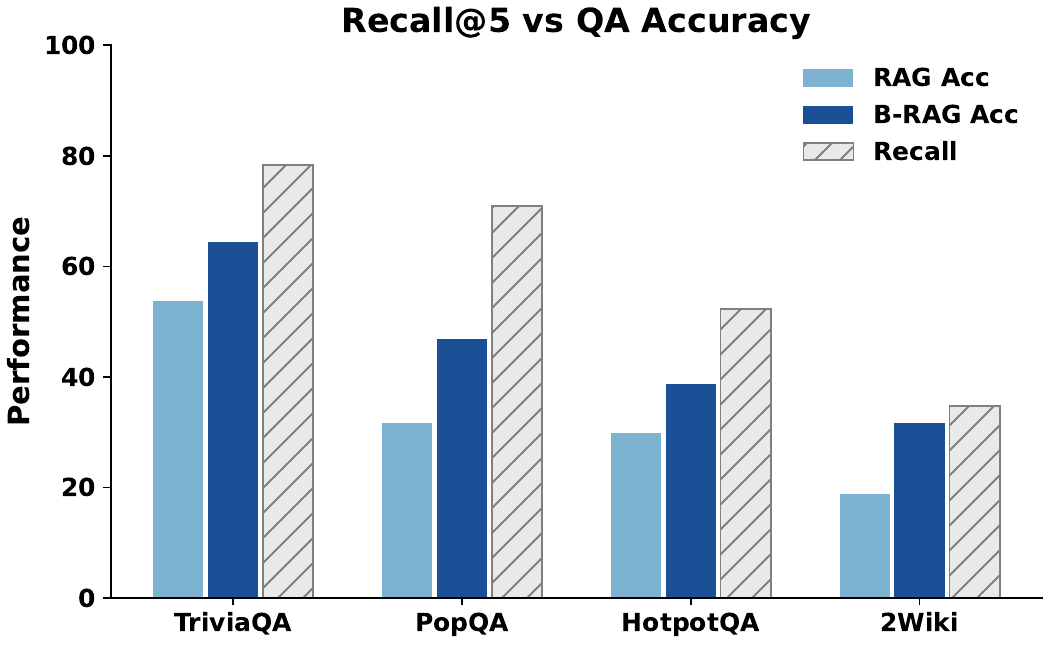}
    \hfill
    \includegraphics[width=0.48\textwidth]{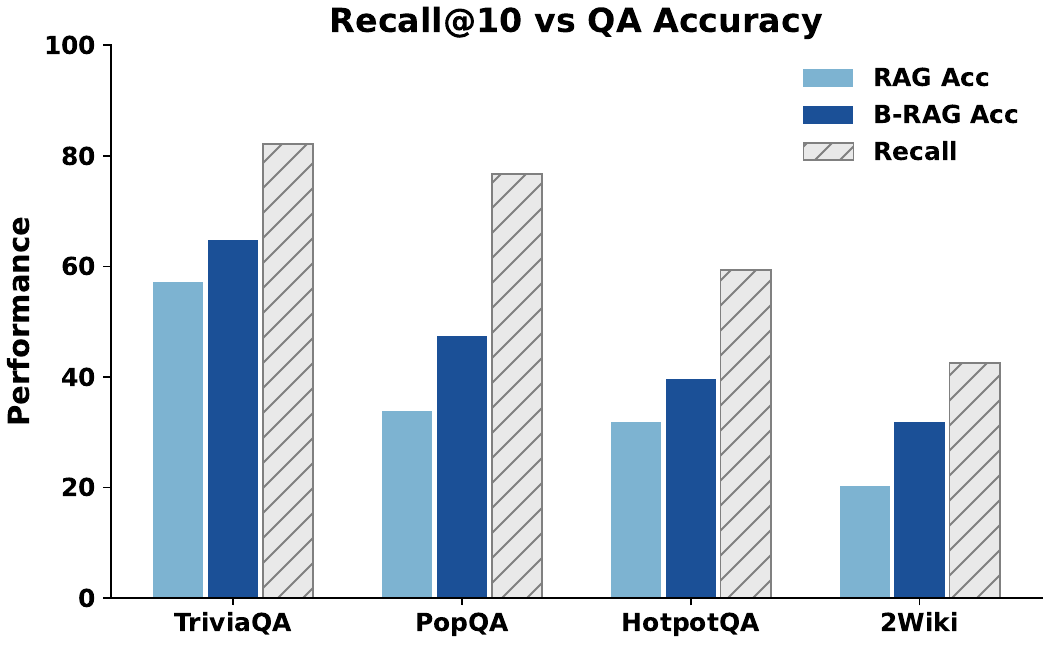}
    \caption{Recall@5 and Recall@10 vs.\ QA Accuracy across different datasets. Higher retrieval recall does not guarantee higher QA accuracy. Our method narrows the gap between recall and accuracy. }
    \label{fig:recall_comparison}
\end{figure*}

Current retrievers optimize exclusively for query-document relevance \citep{e5, qwen3-embedding}, creating two systematic failure modes: they prefer trivial, answer-revealing passages that encourage shortcut learning, and they cannot distinguish genuinely unsolvable evidence from challenging yet sufficient evidence—precisely the kind that best strengthens generator reasoning. Crucially, existing retrievers operate without any estimation of the generator's competence, creating a severe train–test mismatch: systems trained on curated evidence face noisy, incomplete retrieval at deployment, leading to substantial performance degradation \citep{dynamicrag, rankrag}. Empirically, this manifests as a persistent gap between retrieval recall and end-to-end QA accuracy (Figure \ref{fig:recall_comparison}).

% \jh{not sure why your caption is Recall@5/10? should it be Accuracy and Recall@5/10?}
% \sjs{we use Recall@5 and Recall@10 vs.\ QA Accuracy}

To overcome this limitation, we revisit the role of the reranker in RAG systems. Rather than treating it as a passive relevance scorer, we view the reranker as an active evidence set selector, responsible for choosing combinations of documents whose joint structure best supports the generator’s learning and reasoning. Figure \ref{fig:intro} contrasts this paradigm with standard relevance-based reranking. Crucially, evidence sets with similar relevance can induce drastically different learning signals: overly explicit evidence encourages shortcut learning, while incomplete evidence is fundamentally unlearnable. In contrast, evidence that is challenging yet sufficient for inference forces the generator to integrate information and resolve uncertainty.

% \heng{The paragraph after 'Crucially' provides very important information. can you elaborate and walk them through examples, or illustrate them in Figure 1?} \heng{It might help the story flow better if you can give examples that are relevant but not difficult?}
% \sjs{Will add a concrete QA example illustrating the three evidence types (overly explicit / insufficient / challenging-yet-sufficient) and their effects. Will also revise Figure 1 to visualize this contrast. Additionally, will incorporate the "relevant but trivially easy" case as suggested.}

Building on this perspective, we propose \texttt{BAR-RAG}, which instantiates the reranker-as-selector paradigm through boundary-aware evidence design. Rather than maximizing relevance, \texttt{BAR-RAG} explicitly targets the generator’s Goldilocks Zone—evidence sets that are neither trivial nor unsolvable, but lie near the generator’s uncertainty boundary. We operationalize this objective via a two-stage reinforcement learning pipeline: we first train the selector to explore diverse document combinations, guided by rewards based on generator uncertainty and task success, while keeping the generator fixed; we then freeze the selector and fine-tune the generator on the induced evidence distribution, ensuring robustness under noisy retrieval.

% We operationalize this objective via reinforcement learning: the selector explores diverse document combinations, the generator produces candidate answers, and rewards based on generator uncertainty and task success guide the selector toward evidence that is challenging yet solvable. To stabilize this adversarial interaction and align training with deployment conditions, we adopt a two-stage training pipeline in which we first optimize the selector with a fixed generator, and then freeze the selector to fine-tune the generator under the induced evidence distribution, ensuring that the generator is trained on realistic, challenging evidence and improving robustness under noisy retrieval.

We evaluate \texttt{BAR-RAG} on a diverse set of knowledge-intensive question answering benchmarks under realistic retrieval settings. Our results demonstrate that boundary-aware evidence selection consistently improves end-to-end QA performance, yields a significant performance gain of an average 10.3\% over baseline models. Beyond accuracy gains, we show that \texttt{BAR-RAG} reshapes the evidence difficulty distribution toward the generator’s competence boundary, leading to more effective learning signals and substantially improved robustness compared to relevance-based retrieval and reranking baselines.

\section{Method}

\subsection{Overview}

\begin{figure*}
    \centering
    \includegraphics[width=0.98\linewidth]{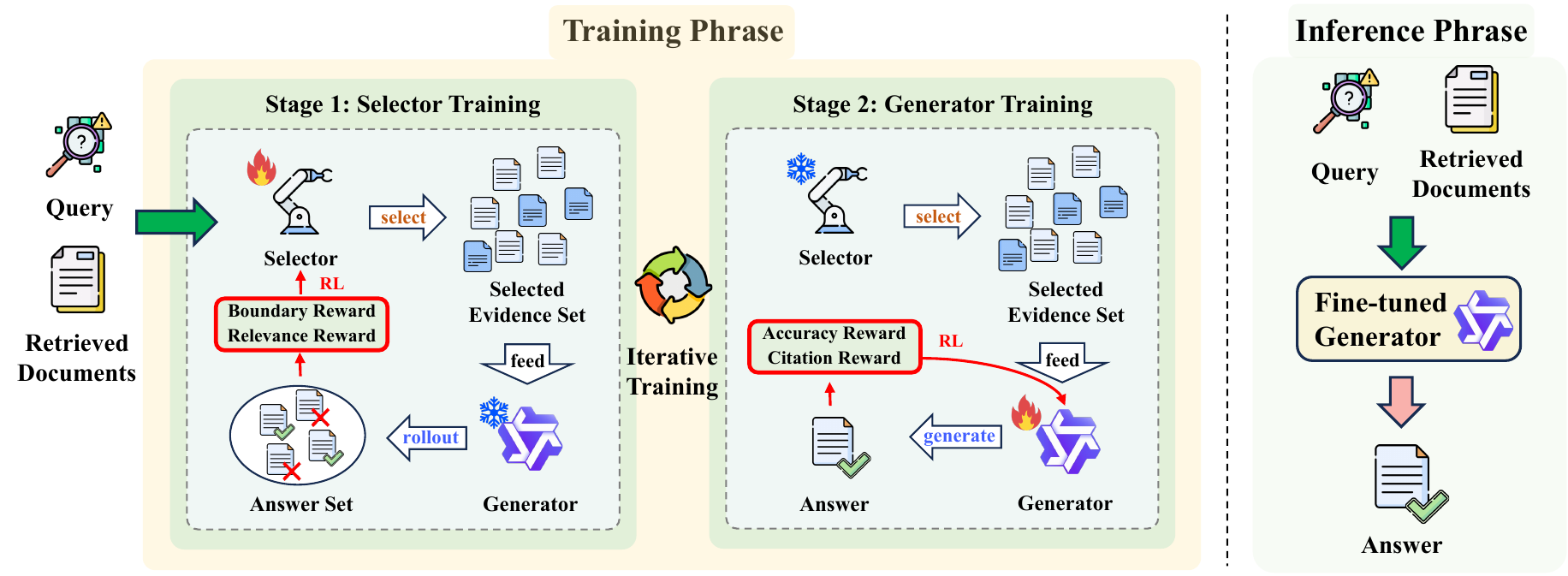}
    \caption{Overview of the \texttt{BAR-RAG} training and inference pipeline. During training, we adopt a two-stage framework: (Stage 1) a selector is trained with reinforcement learning using relevance and uncertainty rewards to identify challenging yet solvable evidence sets, and (Stage 2) the generator is optimized under the induced evidence distribution using accuracy, formatting, and citation rewards. At inference time, the trained generator answers questions using retrieved documents, producing robust and high-quality answers.}
    \label{fig:pipeline}
\end{figure*}

Our approach consists of a two-stage reinforcement learning pipeline (Figure \ref{fig:pipeline}). In Stage~1, we train a selector to identify evidence sets that lie within the generator's ``Goldilocks Zone''---challenging enough to require genuine reasoning, yet solvable given the generator's current competence. In Stage~2, we freeze the selector and fine-tune the generator under the induced evidence distribution, thereby closing the train--test gap that plagues standard RAG systems. We describe each stage below, and introduce an iterative training scheme that progressively refines evidence selection to better match the generator’s evolving competence. We describe each stage below.

\subsection{Problem Setup}

Let $\pi_r$ denote the selector policy and $\pi_g$ the generator policy. Given a query $q$, let $C = \{c_1, \ldots, c_n\}$ be the top-$n$ candidate documents returned by an initial retriever (e.g., BM25 \cite{bm25} or a dense retriever). The selector chooses a subset $S = \{c_{i_1}, \ldots, c_{i_k}\} \subseteq C$ with $k < n$, and the generator produces an answer $a$ conditioned on $(q, S)$.

Our goal is twofold. First, we train the selector policy $\pi_r$ to select evidence sets that maximize the generator’s reasoning performance, not by providing trivial shortcuts, but by identifying evidence that is challenging yet solvable. Second, we train the generator policy $\pi_g$ to produce high-quality, accurate answers conditioned on the evidence sets selected by $\pi_r$, enabling robust reasoning under realistic and challenging retrieval.

\paragraph{Training-time Filtering.}
To ensure informative reinforcement learning signals for selector optimization, we apply a lightweight training-time filtering step that removes queries that are either trivially solvable or fundamentally unanswerable under the retrieved document pool.
The key intuition is that selector learning relies on variance in generator outcomes: queries that are always answered correctly provide no incentive for evidence selection, while queries that are consistently unsolvable yield uniformly low rewards regardless of the selected evidence.
We characterize trivial and unanswerable instances based on the empirical behavior of the generator over multiple rollouts given retrieved documents, using correctness statistics as a proxy for solvability.
This design makes the filtering procedure domain-agnostic and inexpensive, as it depends only on generator feedback rather than task-specific heuristics or additional supervision.
The filtering is applied only during selector training to improve reward signal quality; all queries are retained during evaluation.
Full details are provided in Appendix~\ref{app:filtering}.

\subsection{Stage 1: Boundary-aware Selector Training}

Standard rerankers optimize for relevance alone, often surfacing answer-revealing passages that encourage shortcut learning.
In contrast, we train the selector to target the generator's competence boundary—the region where evidence is neither trivially easy ($\hat{p} \approx 1$) nor impossibly hard ($\hat{p} \approx 0$), but lies near a target difficulty level.

\paragraph{The Goldilocks Zone.}
Let $\hat{p}(S)$ denote the empirical probability that the generator answers correctly given evidence set $S$. We define the Goldilocks Zone as evidence sets satisfying $\hat{p}(S) \approx c$, where $c \in (0, 1)$ is a target correctness rate (e.g., $c = 0.5$). Intuitively, such evidence sets are:
(1) \textbf{Solvable}: The generator can produce correct answers with non-negligible probability, indicating that sufficient information is present.

(2) \textbf{Non-trivial}: The generator does not succeed deterministically, suggesting that reasoning is required rather than simple pattern matching.

By targeting this zone, we encourage the selector to find evidence combinations that demand genuine multi-step reasoning while remaining within the generator's capability.

\paragraph{Selector sampling.}
For each query $q$, we sample $M$ candidate evidence sets from the selector:
\begin{equation}
    S^{(m)} \sim \pi_r(\cdot \mid q, C), \quad m = 1, \ldots, M.
\end{equation}
Each evidence set $S^{(m)}$ is then evaluated using the frozen generator $\pi_g$.
% \jh{It seems $\pi_r$ and $\pi_g$ are both undefined.}
% we defined them at problem setup.

\paragraph{Generator rollouts.}
For each evidence set $S$, we sample $K$ answers from the generator:
\begin{equation}
    a^{(k)} \sim \pi_g(\cdot \mid q, S), \quad k = 1, \ldots, K.
\end{equation}
Each answer must follow a prescribed format (e.g., \verb|<answer>...</answer>|). We define the rollout correctness indicator based on the generator's final reward $R_g(\cdot)$ (defined in Section~\ref{sec:generator_reward_design}).
Specifically, a rollout is considered correct if its reward exceeds a threshold $\delta$:
\begin{equation}
z^{(k)} = \mathbb{I}\left[ R_g\!\left(a^{(k)}\right) \geq \delta \right].
\end{equation}
We then estimate the empirical correctness probability as:
\begin{equation}
\hat{p}(S) = \frac{1}{K} \sum_{k=1}^{K} z^{(k)}.
\end{equation}

\paragraph{Reward design.}
We design four reward components that together guide the selector toward the Goldilocks Zone while maintaining relevance and output quality.

\textbf{(1) Boundary reward $R_{\mathrm{bdy}}(S)$.} This reward encourages evidence sets near the target correctness $c$. We adopt a triangular function that peaks at $\hat{p}(S) = c$ and decreases linearly as the evidence becomes too easy or too hard:
% \begin{equation}
% R_{\mathrm{bdy}}(S) = 1 - \frac{|\hat{p}(S) - c|}{\max(c, 1-c)}.
% \end{equation}
\begin{equation}
R_{\text{bdy}}(S) = \min \left( \frac{\hat{p}(S)}{c}, \frac{1 - \hat{p}(S)}{1 - c} \right)
\end{equation}
To estimate $\hat{p}(S)$, we use the rollout-based definition in the Generator rollouts paragraph with threshold $\delta$. A rollout is deemed correct if its generator reward $R_g(a)$ (defined in Section \ref{sec:generator_reward_design}) exceeds a threshold $\delta$:
\begin{equation}
\hat{p}(S) = \frac{1}{K} \sum_{k=1}^{K} \mathbb{I}\left[ R_g(a^{(k)}) \geq \delta \right].
\end{equation}

\textbf{(2) Relevance reward $R_{\mathrm{rel}}(S)$.} We define the relevance reward as the average retrieval score (in the initial retrieval phrase) of the selected documents,
\begin{equation}
R_{\mathrm{rel}}(S) = \frac{1}{|S|} \sum_{c_i \in S} \mathrm{score}(q, c_i),
\end{equation}
and rescale it to $(0,1)$ for stability.

\textbf{(3) Format reward $R_{\mathrm{fmt}}(S)$.}
We define a binary format indicator that checks whether the selector output is well-formed (e.g., valid document indices, no duplicates, and proper structure):
\begin{equation}
R_{\mathrm{fmt}}(S) = \mathbb{I}\big[\text{$S$ is well-formed}\big].
\end{equation}

This design ensures that malformed selector outputs receive zero reward, while relative trade-offs among boundary targeting, relevance, and document count are only considered for valid evidence sets.

\textbf{(4) Count penalty $P_{\mathrm{cnt}}(S)$.} To encourage the selector to output a target number of documents $k^*$, we apply a penalty proportional to the deviation:
\begin{equation}
P_{\mathrm{cnt}}(S) = \min\left( \alpha \cdot \left| |S| - k^* \right|, \, P_{\max} \right),
\end{equation}
where $\alpha$ is the penalty per document deviation and $P_{\max}$ caps the maximum penalty.

\textbf{Final reward.}
We combine the reward components using the format indicator as a gate:
\begin{equation}
R_r(S) =
R_{\mathrm{fmt}}(S)\cdot
\Big(
\lambda_{\mathrm{bdy}} \, R_{\mathrm{bdy}}(S)
+ \lambda_{\mathrm{rel}} \, R_{\mathrm{rel}}(S)
- P_{\mathrm{cnt}}(S)
\Big),
\end{equation}
where $\lambda_{\mathrm{bdy}}, \lambda_{\mathrm{rel}} \geq 0$ control the relative importance of targeting the competence boundary versus maintaining relevance.

\paragraph{Optimization.}
We optimize the selector using Group Relative Policy Optimization (GRPO) \citep{grpo}. For each group of sampled evidence sets $\{S^{(m)}\}_{m=1}^{M}$, we compute group-normalized advantages:
\begin{equation}
\mathcal{A}^{(m)} = \frac{R_r(S^{(m)}) - \mu}{\sigma + \epsilon},
\end{equation}
where $\mu$ and $\sigma$ are the mean and standard deviation of rewards within the group. The selector is trained to minimize:
\begin{multline}
\mathcal{L}_r(\theta_r) = -\mathbb{E}_{m} \Big[ \min \Big( r^{(m)} \mathcal{A}^{(m)}, \\
\mathrm{clip}(r^{(m)}, 1-\epsilon, 1+\epsilon) \mathcal{A}^{(m)} \Big) \Big],
\end{multline}
where $r^{(m)} = \pi_r(S^{(m)} \mid q, C) / \pi_r^{\mathrm{old}}(S^{(m)} \mid q, C)$ is the likelihood ratio.

\begin{table*}[t]
\centering
\caption{\textbf{\texttt{BAR-RAG} consistently improves performance across models and QA benchmarks.}
Exact match (EM) results on general QA and multi-hop QA tasks for three backbone models.
All values are reported as absolute scores.
For IRCoT, RAG w/ Reranker, RAG SFT, and \texttt{BAR-RAG}, cell shading indicates improvement
(\colorbox{morandiblue!50}{blue}) or degradation (\colorbox{morandired!50}{red}) relative to the RAG baseline.
\texttt{BAR-RAG} yields strong and consistent gains across both single-hop and multi-hop settings,
surpassing standard RAG pipelines and reranking-based baselines.}
\resizebox{0.85\textwidth}{!}{%
\begin{tabular}{lcccccccc}
\toprule
 & \multicolumn{3}{c}{\textbf{General QA}} & \multicolumn{4}{c}{\textbf{Multi-Hop QA}} & \\
\cmidrule(lr){2-4} \cmidrule(lr){5-8}
\textbf{Method} & \textbf{NQ} & \textbf{TriviaQA} & \textbf{PopQA} & \textbf{HotpotQA} & \textbf{2Wiki.} & \textbf{MuSiQue} & \textbf{Bamboogle} & \textbf{Avg.} \\
\midrule
\rowcolor{groupgray}
\multicolumn{9}{c}{\textbf{Qwen-2.5-3B-Instruct}} \\ \midrule
Direct Inference & 10.6 & 28.8 & 10.8 & 14.9 & 24.4 & 2.0 & 2.4 & 13.4 \\
CoT & 2.3 & 3.2 & 0.5 & 2.1 & 2.1 & 0.2 & 0.0 & 1.5 \\ \midrule
RAG & 34.8 & 54.4 & 38.7 & 25.5 & 22.6 & 4.7 & 8.0 & 27.0 \\
IRCoT &
\shadeperf{11.1}{-23.7} & \shadeperf{31.2}{-23.2} & \shadeperf{20.0}{-18.7} & \shadeperf{16.4}{-9.1} &
\shadeperf{17.1}{-5.5} & \shadeperf{6.7}{2.0} & \shadeperf{24.0}{16.0} & \shadeperf{18.1}{-8.9} \\
RAG w/Reranker &
\shadeperf{35.6}{0.8} & \shadeperf{55.3}{0.9} & \shadeperf{39.6}{0.9} & \shadeperf{26.7}{1.2} &
\shadeperf{23.4}{0.8} & \shadeperf{4.9}{0.2} & \shadeperf{10.4}{2.4} & \shadeperf{28.0}{1.0} \\
RAG SFT &
\shadeperf{38.9}{4.1} & \shadeperf{58.1}{3.7} & \shadeperf{41.4}{2.7} & \shadeperf{28.9}{3.4} &
\shadeperf{25.9}{3.3} & \shadeperf{6.4}{1.7} & \shadeperf{13.5}{5.5} & \shadeperf{30.4}{3.4} \\
\texttt{\textbf{BAR-RAG}} (1 Iter) &
\shadeperf{40.1}{5.3} & \shadeperf{58.4}{4.0} & \shadeperf{41.2}{2.5} & \shadeperf{31.0}{5.5} &
\shadeperf{24.9}{2.3} & \shadeperf{6.3}{1.6} & \shadeperf{24.0}{16.0} & \shadeperf{32.3}{5.3} \\
\texttt{\textbf{BAR-RAG}} (2 Iter) &
\shadeperf{41.5}{6.7} & \shadeperf{61.3}{6.9} & \shadeperf{43.4}{4.7} & \shadeperf{33.2}{7.7} &
\shadeperf{26.4}{3.8} & \shadeperf{7.4}{2.7} & \shadeperf{26.3}{18.3} & \shadeperf{34.3}{7.3} \\
\texttt{\textbf{BAR-RAG}} (3 Iter) &
\shadeperf{42.0}{7.2} & \shadeperf{59.7}{5.3} & \shadeperf{44.1}{5.4} & \shadeperf{32.6}{7.1} &
\shadeperf{26.7}{4.1} & \shadeperf{7.2}{2.5} & \shadeperf{26.8}{18.8} & \shadeperf{34.2}{7.2} \\ \midrule

\rowcolor{groupgray}
\multicolumn{9}{c}{\textbf{Qwen-2.5-7B-Instruct}} \\ \midrule
Direct Inference & 13.4 & 40.8 & 14.0 & 18.3 & 12.6 & 3.1 & 12.0 & 16.3 \\
CoT & 4.8 & 18.5 & 5.4 & 9.2 & 10.8 & 2.2 & 23.2 & 10.6 \\ \midrule
RAG & 39.3 & 53.7 & 26.7 & 28.9 & 18.9 & 4.7 & 16.0 & 26.9 \\
IRCoT &
\shadeperf{22.4}{-16.9} & \shadeperf{47.8}{-5.9} & \shadeperf{30.1}{3.4} & \shadeperf{13.3}{-15.6} &
\shadeperf{14.9}{-4.0} & \shadeperf{7.2}{2.5} & \shadeperf{22.4}{6.4} & \shadeperf{22.6}{-4.3} \\
RAG w/Reranker &
\shadeperf{40.5}{1.2} & \shadeperf{55.3}{1.6} & \shadeperf{27.3}{0.6} & \shadeperf{28.1}{-0.8} &
\shadeperf{20.4}{1.5} & \shadeperf{5.5}{0.8} & \shadeperf{18.7}{2.7} & \shadeperf{28.0}{1.1} \\
RAG SFT &
\shadeperf{42.7}{3.4} & \shadeperf{58.6}{4.9} & \shadeperf{32.3}{5.6} & \shadeperf{32.4}{3.5} &
\shadeperf{22.6}{3.7} & \shadeperf{6.8}{2.1} & \shadeperf{27.1}{11.1} & \shadeperf{31.8}{4.9} \\
\texttt{\textbf{BAR-RAG}} (1 Iter) &
\shadeperf{44.7}{5.4} & \shadeperf{63.2}{9.5} & \shadeperf{44.1}{17.4} & \shadeperf{37.2}{8.3} &
\shadeperf{28.3}{9.4} & \shadeperf{8.3}{3.6} & \shadeperf{36.7}{20.7} & \shadeperf{37.2}{10.3} \\
\texttt{\textbf{BAR-RAG}} (2 Iter) &
\shadeperf{46.1}{6.8} & \shadeperf{64.3}{10.6} & \shadeperf{46.3}{19.6} & \shadeperf{38.1}{9.2} &
\shadeperf{29.9}{11.0} & \shadeperf{9.1}{4.4} & \shadeperf{39.1}{23.1} & \shadeperf{38.8}{11.9} \\
\texttt{\textbf{BAR-RAG}} (3 Iter) &
\shadeperf{46.9}{7.6} & \shadeperf{64.5}{10.8} & \shadeperf{46.9}{20.2} & \shadeperf{38.8}{9.9} &
\shadeperf{29.8}{10.9} & \shadeperf{9.1}{4.4} & \shadeperf{39.6}{23.6} & \shadeperf{39.1}{12.2} \\ \midrule

\rowcolor{groupgray}
\multicolumn{9}{c}{\textbf{LLaMA-3.1-8B-Instruct}} \\ \midrule
Direct Inference & 18.4 & 36.5 & 19.8 & 12.5 & 23.0 & 2.7 & 8.8 & 17.4 \\
CoT & 27.8 & 54.1 & 23.5 & 24.1 & 23.0 & 6.8 & 16.3 & 25.1 \\ \midrule
RAG & 42.7 & 58.2 & 28.9 & 30.3 & 19.4 & 6.3 & 17.6 & 29.1 \\
IRCoT &
\shadeperf{23.5}{-19.2} & \shadeperf{48.1}{-10.1} & \shadeperf{31.2}{2.3} & \shadeperf{12.2}{-18.1} &
\shadeperf{11.8}{-7.6} & \shadeperf{6.9}{0.6} & \shadeperf{24.5}{6.9} & \shadeperf{22.6}{-6.5} \\
RAG w/Reranker &
\shadeperf{43.6}{0.9} & \shadeperf{60.1}{1.9} & \shadeperf{30.4}{1.5} & \shadeperf{32.5}{2.2} &
\shadeperf{23.6}{4.2} & \shadeperf{7.6}{1.3} & \shadeperf{19.5}{1.9} & \shadeperf{31.0}{1.9} \\
RAG SFT &
\shadeperf{45.8}{3.1} & \shadeperf{63.4}{5.2} & \shadeperf{35.6}{6.7} & \shadeperf{35.9}{5.6} &
\shadeperf{27.4}{8.0} & \shadeperf{8.9}{2.6} & \shadeperf{24.3}{6.7} & \shadeperf{34.5}{5.4} \\
\texttt{\textbf{BAR-RAG}} (1 Iter) &
\shadeperf{47.5}{4.8} & \shadeperf{65.9}{7.7} & \shadeperf{45.8}{16.9} & \shadeperf{39.5}{9.2} &
\shadeperf{31.4}{12.0} & \shadeperf{11.1}{4.8} & \shadeperf{30.4}{12.8} & \shadeperf{38.8}{9.7} \\
\texttt{\textbf{BAR-RAG}} (2 Iter) &
\shadeperf{49.0}{6.3} & \shadeperf{67.7}{9.5} & \shadeperf{48.0}{19.1} & \shadeperf{40.5}{10.2} &
\shadeperf{33.0}{13.6} & \shadeperf{12.5}{6.2} & \shadeperf{32.8}{15.2} & \shadeperf{40.5}{11.4} \\
\texttt{\textbf{BAR-RAG}} (3 Iter) &
\shadeperf{49.5}{6.8} & \shadeperf{67.3}{9.1} & \shadeperf{48.6}{19.7} & \shadeperf{41.2}{10.9} &
\shadeperf{33.0}{13.6} & \shadeperf{12.0}{5.7} & \shadeperf{33.3}{15.7} & \shadeperf{40.7}{11.6} \\
\bottomrule
\end{tabular}
}
\label{tab:main}
\end{table*}

\subsection{Stage 2: Generator Fine-tuning}

Standard RAG pipelines train generators on curated, near-perfect evidence but deploy them with noisy retrieval, leading to a mismatch and performance degradation. By fine-tuning the generator under the selector's output distribution, we expose it to realistic, challenging evidence during training.

\paragraph{Training procedure.}
For each query $q$, the frozen selector produces an evidence set $S = \pi_r(q, C)$. The generator samples $K$ candidate answers:
\begin{equation}
    a^{(k)} \sim \pi_g(\cdot \mid q, S), \quad k = 1, \ldots, K.
\end{equation}

\paragraph{Generator Reward Design.}
\label{sec:generator_reward_design}
We design a composite reward that encourages both answer accuracy and proper evidence attribution.

\textbf{(1) Format reward.} The generator must produce outputs in the prescribed format (i.e., valid \texttt{<answer>} tags). If the format check fails, the reward is set to zero:
\begin{equation}
R_g(a) = 0 \quad \text{if format is invalid.}
\end{equation}

\textbf{(2) Accuracy reward $R_{\mathrm{acc}}(a)$.} For well-formed outputs, we compute a weighted combination of token-level F1 score and exact match (EM) against the gold answers:
\begin{equation}
R_{\mathrm{acc}}(a) = \beta_1 \cdot \max_{g \in \mathcal{G}} F_1(a, g) + \beta_2 \cdot \max_{g \in \mathcal{G}} \mathrm{EM}(a, g),
\end{equation}
where $\mathcal{G}$ is the set of gold answers, and $\beta_1, \beta_2 \geq 0$ control the relative importance of partial credit (F1) versus exact correctness (EM).

\textbf{(3) Citation reward $R_{\mathrm{cite}}(a)$.} To encourage the generator to ground its reasoning in the provided evidence, we reward appropriate citation behavior in the \texttt{<think>} block. Let $n_{\mathrm{cite}}$ denote the number of unique documents cited. We use a peaked reward centered at a target citation count $n^*$:
\begin{equation}
R_{\mathrm{cite}}(a) = 
\begin{cases}
1.0, & \text{if } n_{\mathrm{cite}} = n^*, \\
0.5, & \text{if } |n_{\mathrm{cite}} - n^*| = 1, \\
0.0, & \text{otherwise}.
\end{cases}
\end{equation}
This encourages the generator to cite a moderate number of sources from the provided documents, sufficient to support multi-hop reasoning but not so many as to dilute focus.

\textbf{Final reward.}
We combine the three components with independent weights for accuracy and citation, while fixing the format coefficient:
\begin{equation}
R_g(a) = R_{\mathrm{fmt}}(a)\cdot \big( \lambda_{\mathrm{acc}} \, R_{\mathrm{acc}}(a) + \lambda_{\mathrm{cite}} \, R_{\mathrm{cite}}(a) \big),
\end{equation}
where $\lambda_{\mathrm{acc}}, \lambda_{\mathrm{cite}} \ge 0$ are hyperparameters.

\paragraph{Optimization.}
The generator is optimized using GRPO with the same clipped objective:
\begin{multline}
\mathcal{L}_g(\theta_g) = -\mathbb{E}_{k} \Big[ \min \Big( r^{(k)} \mathcal{A}^{(k)}, \\
\mathrm{clip}(r^{(k)}, 1-\epsilon, 1+\epsilon) \mathcal{A}^{(k)} \Big) \Big],
\end{multline}
where advantages $\mathcal{A}^{(k)}$ are computed from generator rewards within each group, and $r^{(k)} = \pi_g(a^{(k)} \mid q, S) / \pi_g^{\mathrm{old}}(a^{(k)} \mid q, S)$.

% This generator update completes one iteration of our alternating training cycle.
% By repeatedly re-training the selector to match the generator's current competence boundary and then fine-tuning the generator on the resulting evidence distribution, the system progressively reduces the train--test mismatch and improves robustness under realistic retrieval noise.

\paragraph{Iterative two-stage training.}
Rather than a single pass, we alternate between the two stages for $T$ iterations.
At iteration $t$, we (i) train the selector $\pi_r^{(t)}$ against a frozen generator $\pi_g^{(t-1)}$ to target the current competence boundary, and then
(ii) freeze the updated selector and fine-tune the generator to obtain $\pi_g^{(t)}$ under the induced evidence distribution.
In practice, each iteration consists of one epoch of selector training followed by one epoch of generator fine-tuning.
This alternating procedure progressively refines evidence selection to track the generator's evolving competence. Algorithmic details are provided in Appendix \ref{alg:BAR-RAG}.

\begin{table}[t]
\centering
\caption{Ablation study on different components.}
\label{tab:stage_ablation}
\resizebox{0.45\textwidth}{!}{
\begin{tabular}{lcccc}
\toprule
\textbf{Method} & \textbf{NQ} & \textbf{PopQA} & \textbf{HotpotQA} & \textbf{Avg.} \\
\midrule
Full & 46.9 & 46.9 & 38.8 & 44.2 \\
\midrule
w/o Filtering & 42.6 & 32.1 & 32.3 & 35.6  \\
w/o Stage1 & 42.1 & 42.5 & 41.9 & 42.2 \\
w/o Stage2 & 39.7 & 34.1 & 36.7 & 36.8 \\
\midrule
\textit{Selector} & & & & \\
\quad w/o $R_{\mathrm{bdy}}$ & 43.4 & 43.6 & 33.5 & 40.2 \\
\quad w/o $R_{\mathrm{rel}}$ & 46.1 & 46.3 & 38.1 & 43.5 \\
\textit{Generator} & & & & \\
\quad w/o $R_{\mathrm{cite}}$ & 45.3 & 44.8 & 37.9 & 42.7 \\
\bottomrule
\end{tabular}
}
\end{table}

\subsection{Inference}
The selector is used only during training to shape a challenging evidence distribution for the generator.
At inference time, we discard the selector entirely and apply the fine-tuned generator directly to the top-$k$ documents returned by a standard retriever (e.g., BM25 or a dense retriever), producing the final answer $a = \pi_g(q, S)$.
This design incurs no additional inference cost and preserves the standard RAG pipeline.
By being trained iteratively on adversarially selected evidence near its competence boundary, the generator becomes more robust to noisy, incomplete, and imperfect retrieval results encountered at test time.

\section{Experiments}

\subsection{Datasets and Evaluation Metrics}

We evaluate on seven knowledge-intensive QA datasets spanning diverse reasoning challenges. For \textbf{single-hop QA}, we use Natural Questions (NQ) \cite{nq}, TriviaQA \cite{triviaqa}, and PopQA \cite{popqa}, which test robustness to retrieval noise, paraphrased evidence, and long-tail entity reasoning, respectively. For \textbf{multi-hop QA}, we use HotpotQA (2-hop bridge reasoning) \cite{hotpotqa}, 2WikiMultiHopQA (distant supporting facts) \cite{2wikimqa}, MuSiQue (3--5 compositional hops) \cite{musique}, and Bamboogle (indirect reasoning) \cite{bamboogle}. The training datasets are reported at Appendix \ref{appendix:training_data}. We report Exact Match (EM) as the primary metric, following standard evaluation protocols.

\subsection{Baselines}

We compare against two categories of methods. \textbf{Without retrieval}: (1) Direct Inference, prompting the base model directly; (2) Chain-of-Thought (CoT) \citep{cot}, eliciting step-by-step reasoning; (3) RAG SFT, supervised fine-tuning on QA pairs with retrieved evidence. \textbf{With retrieval}: (1) RAG~\citep{replug}, standard dense retrieval-augmented generation; (2) RAG w/ Reranker, adding a neural reranker to re-score retrieved documents before generation; (3) IRCoT \cite{IRCoT}, a multi-step QA framework that interleaves retrieval with steps in a CoT;  (4) RAG SFT, fine-tuning on QA pairs augmented with top-5 retrieved passages, exposing the model to both relevant and noisy evidence during training. All baselines use the same base model (\textit{Qwen2.5-3B-Instruct}, \textit{Qwen2.5-7B-Instruct} \cite{qwen25} and \textit{LLaMA-3.1-8B-Instruct} \cite{llama3}), retriever (\textit{E5-base-v2} \cite{e5}) and reranker (\textit{Qwen-3-Embedding-8B} \cite{qwen3-embedding}) and Top-5 retrieved documents as input for fair comparison.

\subsection{Implementation Details}

We use instruction-tuned LLMs as both the selector and generator backbones, including Qwen2.5 and LLaMA-3.1 variants.
Detailed model configurations, reward parameters, rollout settings, and training schedules are provided in Appendix~\ref{sec:impl_details}.

\subsection{Main Results}
Table \ref{tab:main} presents our main results across seven QA benchmarks. \texttt{BAR-RAG} consistently outperforms all baselines on both general and multi-hop QA tasks. Using \textit{Qwen2.5-7B-Instruct} as a representative example, \texttt{BAR-RAG} achieves substantial improvements over the strongest baseline RAG SFT on single-hop benchmarks: +8.5 on NQ (51.2 vs.\ 42.7), +5.9 on TriviaQA, and +14.6 on PopQA. The advantages are more pronounced on multi-hop tasks, with gains of +6.4 on HotpotQA, +7.2 on 2WikiMultiHopQA, and +12.5 on Bamboogle—the latter highlighting the benefit of training on evidence that demands genuine multi-step reasoning. Results on \textit{LLaMA-3.1-8B-Instruct} and \textit{Qwen2.5-3B-Instruct} confirm that our approach generalizes across model families, achieving average EM of 40.7 and 34.2 respectively (vs.\ 29.1 and 27.0 for RAG).

\begin{figure*}[t]
    \centering

    % ---------------- Row 1 ----------------
    \begin{minipage}{0.24\textwidth}
        \centering
        \includegraphics[width=\linewidth]{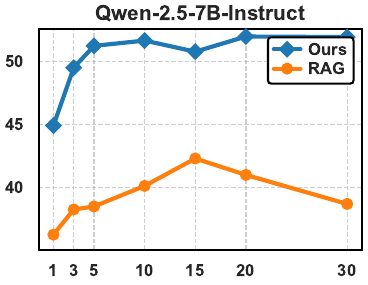}
    \end{minipage}\hfill
    \begin{minipage}{0.24\textwidth}
        \centering
        \includegraphics[width=\linewidth]{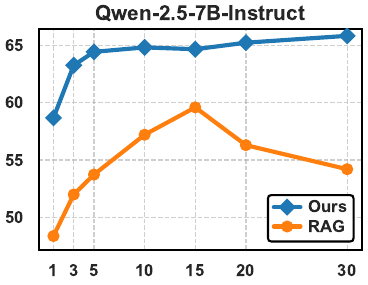}
    \end{minipage}\hfill
    \begin{minipage}{0.24\textwidth}
        \centering
        \includegraphics[width=\linewidth]{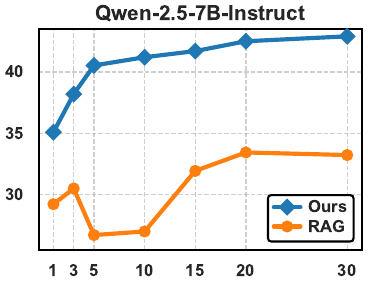}
    \end{minipage}\hfill
    \begin{minipage}{0.24\textwidth}
        \centering
        \includegraphics[width=\linewidth]{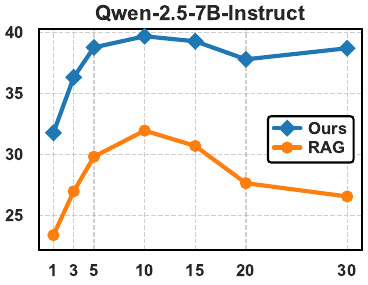}
    \end{minipage}

    \vspace{0.35em}

    % ---------------- Row 2 ----------------
    \begin{minipage}{0.24\textwidth}
        \centering
        \includegraphics[width=\linewidth]{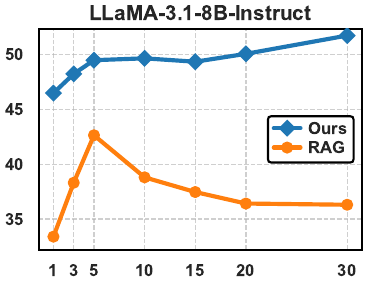}
    \end{minipage}\hfill
    \begin{minipage}{0.24\textwidth}
        \centering
        \includegraphics[width=\linewidth]{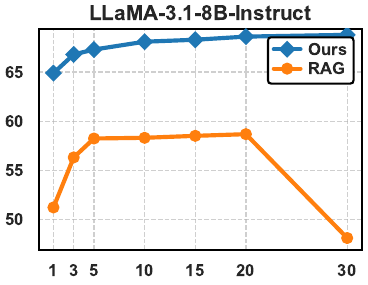}
    \end{minipage}\hfill
    \begin{minipage}{0.24\textwidth}
        \centering
        \includegraphics[width=\linewidth]{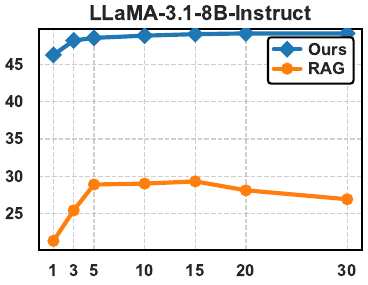}
    \end{minipage}\hfill
    \begin{minipage}{0.24\textwidth}
        \centering
        \includegraphics[width=\linewidth]{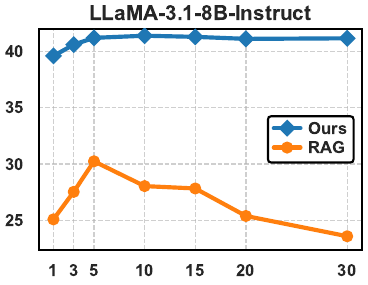}
    \end{minipage}

    \caption{
Top-$K$ accuracy curves on four QA benchmarks for two base models: \textit{Qwen-2.5-7B-Instruct} and \textit{LLaMA-3.1-8B-Instruct}. From left to right, columns correspond to \textbf{NQ}, \textbf{TriviaQA}, \textbf{PopQA}, and \textbf{HotpotQA}. Across both models and all datasets, our method consistently achieves higher accuracy in low-$K$ regimes and remains robust as $K$ increases, whereas standard RAG exhibits weaker scaling behavior and higher sensitivity to retrieval noise.
}
    \label{fig:robustness_k}
\end{figure*}

\paragraph{Iterative Improvement}
We examine how performance evolves across training iterations. As shown in Table \ref{tab:main}, all three models exhibit consistent improvement from Iter 1 to Iter 2, with diminishing gains from Iter 2 to Iter 3. This pattern suggests that the iterative co-training procedure converges toward a stable solution within a few iterations, and that the majority of gains are captured in the first two rounds of selector-generator co-adaptation.

\subsection{Ablation Studies}
\label{sec:ablation}

We conduct ablation studies to validate the necessity of key design choices, including (i) the design of reward components for boundary-aware evidence selection and (ii) the two-stage training pipeline.

\begin{figure*}[t]
    \centering
    \includegraphics[width=0.48\textwidth]{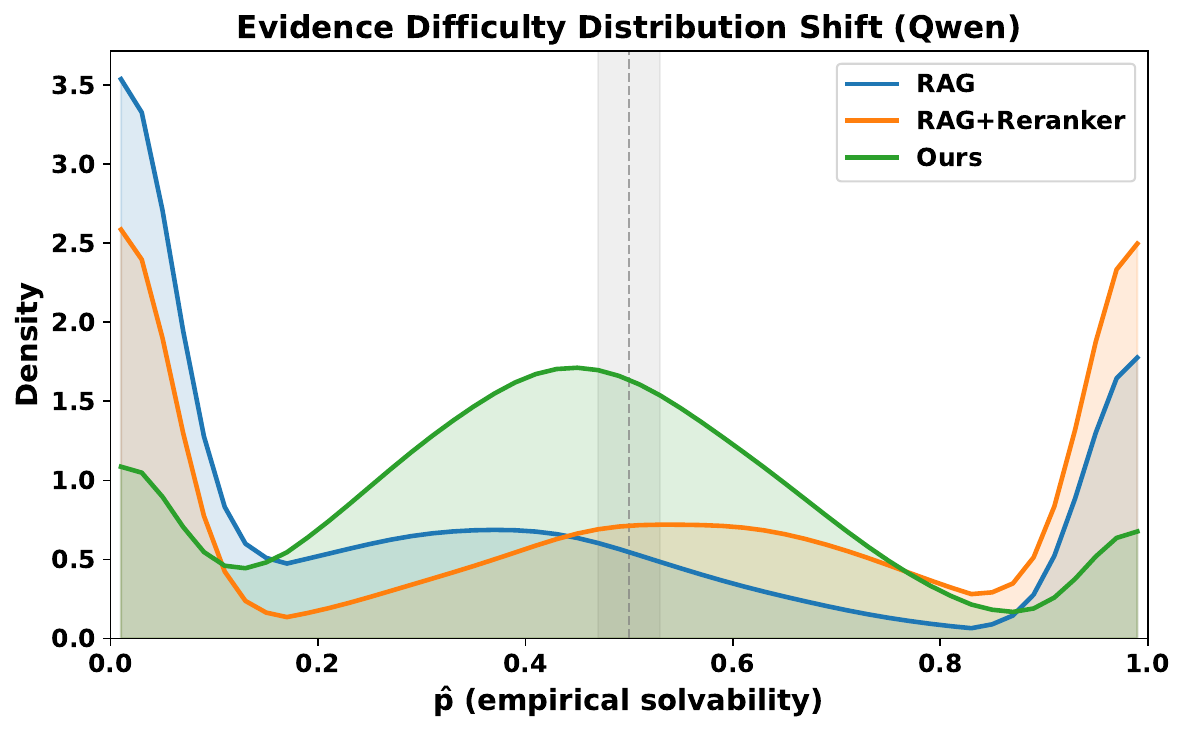}
    \hfill
    \includegraphics[width=0.48\textwidth]{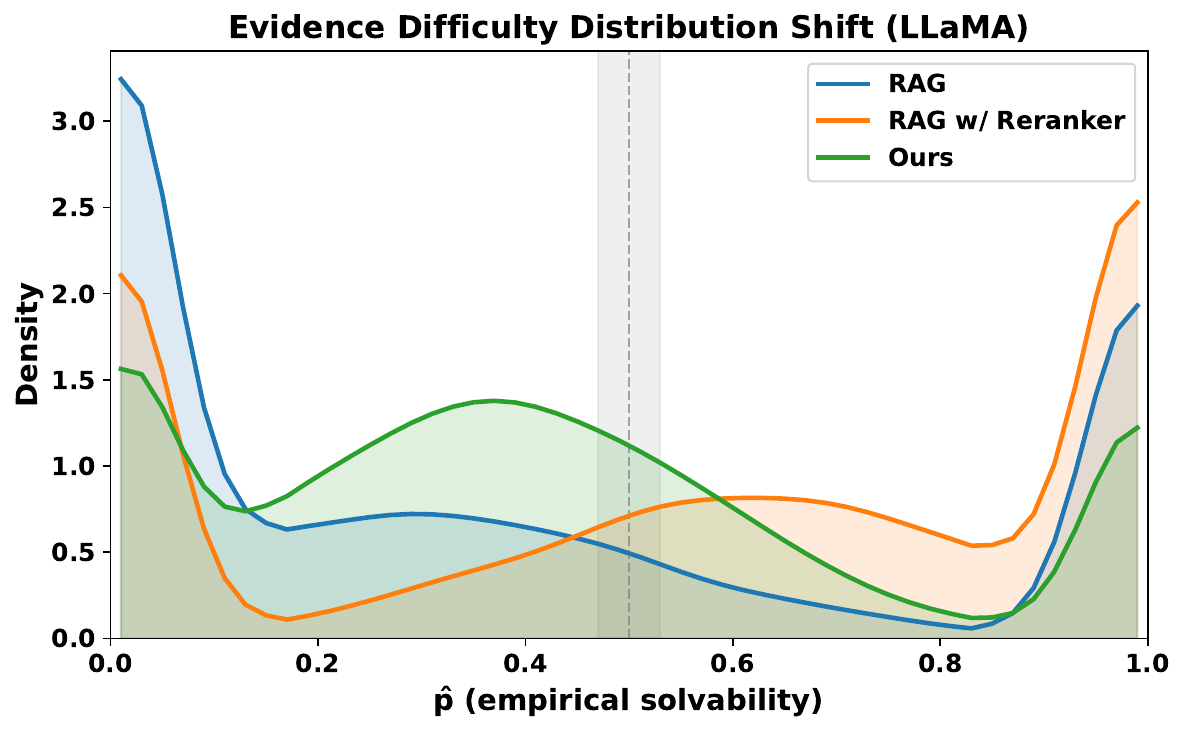}
    \caption{Distribution shift of evidence difficulty ($\hat{p}$) across different retrieval methods. Our method concentrates samples toward the target correctness level $c{=}0.5$, substantially reducing extreme cases near 0 (unsolvable) and 1 (trivially solvable).}
    \label{fig:dis_shift}
\end{figure*}

\paragraph{Effect of Reward Components.}
To examine the contribution of individual reward terms in the selector, we ablate two key components: the boundary reward $R_{\mathrm{bdy}}$ and the relevance reward $R_{\mathrm{rel}}$, while keeping other training settings unchanged.
Specifically, we evaluate variants that remove $R_{\mathrm{bdy}}$ or $R_{\mathrm{rel}}$ from the selector training objective.
Results are reported in Table~\ref{tab:stage_ablation}.
Removing the boundary reward leads to a clear performance degradation across all datasets, confirming that targeting evidence near the generator's competence boundary is crucial for robust reasoning.
In contrast, removing the relevance reward results in a milder but consistent drop, indicating its complementary role in maintaining evidence quality.
We further ablate the citation reward $R_{\mathrm{cite}}$ used in generator fine-tuning. Removing $R_{\mathrm{cite}}$ results in a consistent but smaller performance drop, indicating that citation supervision complements competence-aware evidence selection by improving answer grounding rather than driving the main gains.

\paragraph{Effect of Two-Stage Training.}
Our pipeline consists of two stages: selector training (Stage~1) followed by generator fine-tuning (Stage~2).
To assess the necessity of each component, we conduct ablation studies summarized in Table~\ref{tab:stage_ablation}.
When Stage~2 is removed, the selector is trained while keeping the generator frozen at its initial state.
This variant leads to a clear performance drop across all benchmarks, indicating that adapting the generator to the competence-boundary evidence distribution is crucial for effective reasoning.
We further examine the role of the training-time filtering step by removing it and training the selector on the full dataset.
Eliminating filtering consistently degrades performance, with particularly large drops on PopQA and HotpotQA, confirming that trivially solvable and unanswerable queries introduce degenerate RL signals that hinder stable selector optimization.
Finally, removing Stage~1 eliminates competence-aware evidence selection altogether, reducing the pipeline to standard retrieval followed by generator training.
The resulting degradation demonstrates that naive retrieval is insufficient for robust multi-hop reasoning.

\subsection{Analysis}
\label{sec:analysis}

% We provide analyses to understand where performance gains come from and why \texttt{BAR-RAG} improves robustness.

\paragraph{Generator Robustness to Evidence Quality.}
A central claim of our approach is that training on competence-boundary evidence yields a generator that is inherently more robust to evidence quality—even when the selector is discarded at inference time.
To verify this, we evaluate generators trained with \texttt{BAR-RAG}, and standard RAG, using the same naive retriever outputs (top-$k$ by retrieval score) under varying evidence budgets $k \in \{1, 3, 5, 10, 15, 30\}$.
We conduct this evaluation across two backbone models (\textit{Qwen-2.5-7B-Instruct} and \textit{LLaMA-3.1-8B-Instruct}) and four datasets (NQ, TriviaQA, PopQA, and HotpotQA).
Figure~\ref{fig:robustness_k} reports K-accuracy curves for all settings.
Consistent with our hypothesis, \texttt{BAR-RAG}-trained generators consistently outperform baselines across all $k$, with the largest gains appearing in low-$k$ regimes where evidence is sparse and reasoning robustness is most critical.

\paragraph{Evidence Difficulty Distribution Shift.}
For each question, we estimate the empirical solvability $\hat{p}(S)$ of an evidence set $S$ via $K$ generator rollouts, and compare the resulting distributions across three methods: Naive RAG (selecting top-$k$ documents by retrieval score), RAG with a neural reranker, and our \texttt{BAR-RAG} selector.
Figure~\ref{fig:dis_shift} compares the resulting distributions across Naive RAG, RAG with a neural reranker, and our \texttt{BAR-RAG} selector for both \textit{Qwen-2.5-7B-Instruct} and \textit{LLaMA-3.1-8B-Instruct}.
While Naive RAG and reranker-based RAG exhibit bimodal distributions dominated by unsolvable or trivially answer-revealing contexts, \texttt{BAR-RAG} suppresses both extremes and concentrates probability mass around the target correctness level $c=0.5$.
This consistent shift across generators indicates that \texttt{BAR-RAG} actively promotes hard-but-solvable evidence near the generator’s competence boundary rather than merely improving relevance.

Additional analyses, including counterfactual evidence-dependence studies and comparisons with agentic RAG methods, are provided in Appendix \ref{app:counterfactual} and \ref{app:agentic_comparison}.

% \paragraph{Training Efficiency.} We compare the training efficiency of \texttt{BAR-RAG} against RAG SFT by plotting downstream QA performance (average EM) against the number of training steps. Figure~\ref{fig:efficiency} shows the learning curves. This analysis reveals whether \texttt{BAR-RAG} achieves higher final performance, faster convergence, or both compared to standard supervised fine-tuning on retrieved evidence.

\section{Related Work}

% \jh{Related work review seems to be too short and too superficial? More references and some concrete evidence to support your claim on new methods should be included.}
% \sjs{due to space limit, we put more RW to Appendix.}
\paragraph{Retrieval-Augmented Generation}
Retrieval-Augmented Generation (RAG) improves factual accuracy by grounding language model outputs in external knowledge sources \cite{lewis2021retrievalaugmentedgenerationknowledgeintensivenlp, Guu2020RAG}.
Recent work explores tighter retrieval--generation integration, including mechanisms that adapt retrieval behavior based on model uncertainty or self-reflection, such as Self-RAG \cite{selfrag}.
However, \texttt{BAR-RAG} targets what evidence to retrieve, framing evidence selection as a competence-aware process that deliberately selects hard-but-solvable contexts to strengthen the generator’s reasoning ability.

\paragraph{Document Reranking}
Reranking refines retrieved candidates before downstream generation, with early neural approaches relying on cross-encoder architectures to compute relevance scores \cite{nogueira2019passage}.
More recently, reinforcement learning has been used to optimize reranking policies via generator feedback, as in DynamicRAG \cite{dynamicrag}.
In contrast to prior RL-based rerankers that optimize relevance or answer quality, \texttt{BAR-RAG} introduces an explicit competence boundary reward that directly models evidence difficulty, enabling the selection of challenging yet solvable evidence sets.

Further extand related works are reported in Appendix \ref{sec:extended_rw}.

\section{Conclusion}
We presented \texttt{BAR-RAG}, a boundary-aware evidence selection training framework that trains the selector to target hard-but-solvable evidence near the generator's competence boundary. We train generator with realistic retrieval conditions, yielding consistent robustness gains across diverse QA benchmarks without additional inference-time cost.

% \section*{Impact Statement}

% This paper presents work whose goal is to advance the field of machine learning. There are many potential societal consequences of our work, none of which we feel must be specifically highlighted here.

\bibliography{example_paper}

@article{dynamicrag,
  author       = {Jiashuo Sun and
                  Xianrui Zhong and
                  Sizhe Zhou and
                  Jiawei Han},
  title        = {DynamicRAG: Leveraging Outputs of Large Language Model as Feedback
                  for Dynamic Reranking in Retrieval-Augmented Generation},
  journal      = {CoRR},
  volume       = {abs/2505.07233},
  year         = {2025},
  url          = {https://doi.org/10.48550/arXiv.2505.07233},
  doi          = {10.48550/ARXIV.2505.07233},
  eprinttype    = {arXiv},
  eprint       = {2505.07233},
}

@article{agentic_ai,
  title={Adaptation of Agentic AI},
  author={Jiang, Pengcheng and Lin, Jiacheng and Shi, Zhiyi and Wang, Zifeng and He, Luxi and Wu, Yichen and Zhong, Ming and Song, Peiyang and Zhang, Qizheng and Wang, Heng and others},
  journal={arXiv preprint arXiv:2512.16301},
  year={2025}
}

@article{rag_survey,
  author       = {Yangning Li and
                  Weizhi Zhang and
                  Yuyao Yang and
                  Wei{-}Chieh Huang and
                  Yaozu Wu and
                  Junyu Luo and
                  Yuanchen Bei and
                  Henry Peng Zou and
                  Xiao Luo and
                  Yusheng Zhao and
                  Chunkit Chan and
                  Yankai Chen and
                  Zhongfen Deng and
                  Yinghui Li and
                  Hai{-}Tao Zheng and
                  Dongyuan Li and
                  Renhe Jiang and
                  Ming Zhang and
                  Yangqiu Song and
                  Philip S. Yu},
  title        = {Towards Agentic {RAG} with Deep Reasoning: {A} Survey of RAG-Reasoning
                  Systems in LLMs},
  journal      = {CoRR},
  volume       = {abs/2507.09477},
  year         = {2025},
  url          = {https://doi.org/10.48550/arXiv.2507.09477},
  doi          = {10.48550/ARXIV.2507.09477},
  eprinttype    = {arXiv},
  eprint       = {2507.09477},
}

@article{e5,
  author       = {Liang Wang and
                  Nan Yang and
                  Xiaolong Huang and
                  Binxing Jiao and
                  Linjun Yang and
                  Daxin Jiang and
                  Rangan Majumder and
                  Furu Wei},
  title        = {Text Embeddings by Weakly-Supervised Contrastive Pre-training},
  journal      = {CoRR},
  volume       = {abs/2212.03533},
  year         = {2022},
  url          = {https://doi.org/10.48550/arXiv.2212.03533},
  doi          = {10.48550/ARXIV.2212.03533},
  eprinttype    = {arXiv},
  eprint       = {2212.03533},
}

@article{qwen3-embedding,
  author       = {Yanzhao Zhang and
                  Mingxin Li and
                  Dingkun Long and
                  Xin Zhang and
                  Huan Lin and
                  Baosong Yang and
                  Pengjun Xie and
                  An Yang and
                  Dayiheng Liu and
                  Junyang Lin and
                  Fei Huang and
                  Jingren Zhou},
  title        = {Qwen3 Embedding: Advancing Text Embedding and Reranking Through Foundation
                  Models},
  journal      = {CoRR},
  volume       = {abs/2506.05176},
  year         = {2025},
  url          = {https://doi.org/10.48550/arXiv.2506.05176},
  doi          = {10.48550/ARXIV.2506.05176},
  eprinttype    = {arXiv},
  eprint       = {2506.05176},
}

@article{nq,
    title = "Natural Questions: A Benchmark for Question Answering Research",
    author = "Kwiatkowski, Tom  and
      Palomaki, Jennimaria  and
      Redfield, Olivia  and
      Collins, Michael  and
      Parikh, Ankur  and
      Alberti, Chris  and
      Epstein, Danielle  and
      Polosukhin, Illia  and
      Devlin, Jacob  and
      Lee, Kenton  and
      Toutanova, Kristina  and
      Jones, Llion  and
      Kelcey, Matthew  and
      Chang, Ming-Wei  and
      Dai, Andrew M.  and
      Uszkoreit, Jakob  and
      Le, Quoc  and
      Petrov, Slav",
    editor = "Lee, Lillian  and
      Johnson, Mark  and
      Roark, Brian  and
      Nenkova, Ani",
    journal = "Transactions of the Association for Computational Linguistics",
    volume = "7",
    year = "2019",
    address = "Cambridge, MA",
    publisher = "MIT Press",
    url = "https://aclanthology.org/Q19-1026/",
    doi = "10.1162/tacl_a_00276",
    pages = "452--466",
}

@inproceedings{triviaqa,
    title = "{T}rivia{QA}: A Large Scale Distantly Supervised Challenge Dataset for Reading Comprehension",
    author = "Joshi, Mandar  and
      Choi, Eunsol  and
      Weld, Daniel  and
      Zettlemoyer, Luke",
    editor = "Barzilay, Regina  and
      Kan, Min-Yen",
    booktitle = "Proceedings of the 55th Annual Meeting of the Association for Computational Linguistics (Volume 1: Long Papers)",
    month = jul,
    year = "2017",
    address = "Vancouver, Canada",
    publisher = "Association for Computational Linguistics",
    url = "https://aclanthology.org/P17-1147/",
    doi = "10.18653/v1/P17-1147",
    pages = "1601--1611",
}

@inproceedings{popqa,
  author       = {Alex Mallen and
                  Akari Asai and
                  Victor Zhong and
                  Rajarshi Das and
                  Daniel Khashabi and
                  Hannaneh Hajishirzi},
  editor       = {Anna Rogers and
                  Jordan L. Boyd{-}Graber and
                  Naoaki Okazaki},
  title        = {When Not to Trust Language Models: Investigating Effectiveness of
                  Parametric and Non-Parametric Memories},
  booktitle    = {Proceedings of the 61st Annual Meeting of the Association for Computational
                  Linguistics (Volume 1: Long Papers), {ACL} 2023, Toronto, Canada,
                  July 9-14, 2023},
  pages        = {9802--9822},
  publisher    = {Association for Computational Linguistics},
  year         = {2023},
  url          = {https://doi.org/10.18653/v1/2023.acl-long.546},
  doi          = {10.18653/V1/2023.ACL-LONG.546},
}

@inproceedings{hotpotqa,
  author       = {Zhilin Yang and
                  Peng Qi and
                  Saizheng Zhang and
                  Yoshua Bengio and
                  William W. Cohen and
                  Ruslan Salakhutdinov and
                  Christopher D. Manning},
  editor       = {Ellen Riloff and
                  David Chiang and
                  Julia Hockenmaier and
                  Jun'ichi Tsujii},
  title        = {HotpotQA: {A} Dataset for Diverse, Explainable Multi-hop Question
                  Answering},
  booktitle    = {Proceedings of the 2018 Conference on Empirical Methods in Natural
                  Language Processing, Brussels, Belgium, October 31 - November 4, 2018},
  pages        = {2369--2380},
  publisher    = {Association for Computational Linguistics},
  year         = {2018},
  url          = {https://doi.org/10.18653/v1/d18-1259},
  doi          = {10.18653/V1/D18-1259},
}

@inproceedings{2wikimqa,
    title = "Constructing A Multi-hop {QA} Dataset for Comprehensive Evaluation of Reasoning Steps",
    author = "Ho, Xanh  and
      Duong Nguyen, Anh-Khoa  and
      Sugawara, Saku  and
      Aizawa, Akiko",
    editor = "Scott, Donia  and
      Bel, Nuria  and
      Zong, Chengqing",
    booktitle = "Proceedings of the 28th International Conference on Computational Linguistics",
    month = dec,
    year = "2020",
    address = "Barcelona, Spain (Online)",
    publisher = "International Committee on Computational Linguistics",
    url = "https://aclanthology.org/2020.coling-main.580/",
    doi = "10.18653/v1/2020.coling-main.580",
    pages = "6609--6625",
}

@article{musique,
    title = "♫ {M}u{S}i{Q}ue: Multihop Questions via Single-hop Question Composition",
    author = "Trivedi, Harsh  and
      Balasubramanian, Niranjan  and
      Khot, Tushar  and
      Sabharwal, Ashish",
    editor = "Roark, Brian  and
      Nenkova, Ani",
    journal = "Transactions of the Association for Computational Linguistics",
    volume = "10",
    year = "2022",
    address = "Cambridge, MA",
    publisher = "MIT Press",
    url = "https://aclanthology.org/2022.tacl-1.31/",
    doi = "10.1162/tacl_a_00475",
    pages = "539--554",
}

@inproceedings{bamboogle,
  author       = {Ofir Press and
                  Muru Zhang and
                  Sewon Min and
                  Ludwig Schmidt and
                  Noah A. Smith and
                  Mike Lewis},
  editor       = {Houda Bouamor and
                  Juan Pino and
                  Kalika Bali},
  title        = {Measuring and Narrowing the Compositionality Gap in Language Models},
  booktitle    = {Findings of the Association for Computational Linguistics: {EMNLP}
                  2023, Singapore, December 6-10, 2023},
  pages        = {5687--5711},
  publisher    = {Association for Computational Linguistics},
  year         = {2023},
  url          = {https://doi.org/10.18653/v1/2023.findings-emnlp.378},
  doi          = {10.18653/V1/2023.FINDINGS-EMNLP.378},
}

@inproceedings{cot,
  author       = {Jason Wei and
                  Xuezhi Wang and
                  Dale Schuurmans and
                  Maarten Bosma and
                  Brian Ichter and
                  Fei Xia and
                  Ed H. Chi and
                  Quoc V. Le and
                  Denny Zhou},
  editor       = {Sanmi Koyejo and
                  S. Mohamed and
                  A. Agarwal and
                  Danielle Belgrave and
                  K. Cho and
                  A. Oh},
  title        = {Chain-of-Thought Prompting Elicits Reasoning in Large Language Models},
  booktitle    = {Advances in Neural Information Processing Systems 35: Annual Conference
                  on Neural Information Processing Systems 2022, NeurIPS 2022, New Orleans,
                  LA, USA, November 28 - December 9, 2022},
  year         = {2022},
  url          = {http://papers.nips.cc/paper\_files/paper/2022/hash/9d5609613524ecf4f15af0f7b31abca4-Abstract-Conference.html},
}

@inproceedings{power_of_noise,
  author       = {Florin Cuconasu and
                  Giovanni Trappolini and
                  Federico Siciliano and
                  Simone Filice and
                  Cesare Campagnano and
                  Yoelle Maarek and
                  Nicola Tonellotto and
                  Fabrizio Silvestri},
  editor       = {Grace Hui Yang and
                  Hongning Wang and
                  Sam Han and
                  Claudia Hauff and
                  Guido Zuccon and
                  Yi Zhang},
  title        = {The Power of Noise: Redefining Retrieval for {RAG} Systems},
  booktitle    = {Proceedings of the 47th International {ACM} {SIGIR} Conference on
                  Research and Development in Information Retrieval, {SIGIR} 2024, Washington
                  DC, USA, July 14-18, 2024},
  pages        = {719--729},
  publisher    = {{ACM}},
  year         = {2024},
  url          = {https://doi.org/10.1145/3626772.3657834},
  doi          = {10.1145/3626772.3657834},
}

@inproceedings{rankrag,
  author       = {Yue Yu and
                  Wei Ping and
                  Zihan Liu and
                  Boxin Wang and
                  Jiaxuan You and
                  Chao Zhang and
                  Mohammad Shoeybi and
                  Bryan Catanzaro},
  editor       = {Amir Globersons and
                  Lester Mackey and
                  Danielle Belgrave and
                  Angela Fan and
                  Ulrich Paquet and
                  Jakub M. Tomczak and
                  Cheng Zhang},
  title        = {RankRAG: Unifying Context Ranking with Retrieval-Augmented Generation
                  in LLMs},
  booktitle    = {Advances in Neural Information Processing Systems 38: Annual Conference
                  on Neural Information Processing Systems 2024, NeurIPS 2024, Vancouver,
                  BC, Canada, December 10 - 15, 2024},
  year         = {2024},
  url          = {http://papers.nips.cc/paper\_files/paper/2024/hash/db93ccb6cf392f352570dd5af0a223d3-Abstract-Conference.html},
}

@misc{ragged,
      title={RAGGED: Towards Informed Design of Scalable and Stable RAG Systems}, 
      author={Jennifer Hsia and Afreen Shaikh and Zhiruo Wang and Graham Neubig},
      year={2025},
      eprint={2403.09040},
      archivePrefix={arXiv},
      primaryClass={cs.CL},
      url={https://arxiv.org/abs/2403.09040}, 
}

@article{grpo,
  author       = {Zhihong Shao and
                  Peiyi Wang and
                  Qihao Zhu and
                  Runxin Xu and
                  Junxiao Song and
                  Mingchuan Zhang and
                  Y. K. Li and
                  Y. Wu and
                  Daya Guo},
  title        = {DeepSeekMath: Pushing the Limits of Mathematical Reasoning in Open
                  Language Models},
  journal      = {CoRR},
  volume       = {abs/2402.03300},
  year         = {2024},
  url          = {https://doi.org/10.48550/arXiv.2402.03300},
  doi          = {10.48550/ARXIV.2402.03300},
  eprinttype    = {arXiv},
  eprint       = {2402.03300},
}

@article{bm25,
  author       = {Stephen E. Robertson and
                  Hugo Zaragoza},
  title        = {The Probabilistic Relevance Framework: {BM25} and Beyond},
  journal      = {Found. Trends Inf. Retr.},
  volume       = {3},
  number       = {4},
  pages        = {333--389},
  year         = {2009},
  url          = {https://doi.org/10.1561/1500000019},
  doi          = {10.1561/1500000019},
}

@inproceedings{replug,
    title = "{REPLUG}: Retrieval-Augmented Black-Box Language Models",
    author = "Shi, Weijia  and
      Min, Sewon  and
      Yasunaga, Michihiro  and
      Seo, Minjoon  and
      James, Richard  and
      Lewis, Mike  and
      Zettlemoyer, Luke  and
      Yih, Wen-tau",
    editor = "Duh, Kevin  and
      Gomez, Helena  and
      Bethard, Steven",
    booktitle = "Proceedings of the 2024 Conference of the North American Chapter of the Association for Computational Linguistics: Human Language Technologies (Volume 1: Long Papers)",
    month = jun,
    year = "2024",
    address = "Mexico City, Mexico",
    publisher = "Association for Computational Linguistics",
    url = "https://aclanthology.org/2024.naacl-long.463/",
    doi = "10.18653/v1/2024.naacl-long.463",
    pages = "8371--8384",
}

@misc{qwen25,
      title={Qwen2.5 Technical Report}, 
      author={Qwen and : and An Yang and Baosong Yang and Beichen Zhang and Binyuan Hui and Bo Zheng and Bowen Yu and Chengyuan Li and Dayiheng Liu and Fei Huang and Haoran Wei and Huan Lin and Jian Yang and Jianhong Tu and Jianwei Zhang and Jianxin Yang and Jiaxi Yang and Jingren Zhou and Junyang Lin and Kai Dang and Keming Lu and Keqin Bao and Kexin Yang and Le Yu and Mei Li and Mingfeng Xue and Pei Zhang and Qin Zhu and Rui Men and Runji Lin and Tianhao Li and Tianyi Tang and Tingyu Xia and Xingzhang Ren and Xuancheng Ren and Yang Fan and Yang Su and Yichang Zhang and Yu Wan and Yuqiong Liu and Zeyu Cui and Zhenru Zhang and Zihan Qiu},
      year={2025},
      eprint={2412.15115},
      archivePrefix={arXiv},
      primaryClass={cs.CL},
      url={https://arxiv.org/abs/2412.15115}, 
}

@misc{llama3,
  title         = {The Llama 3 Herd of Models},
  author        = {Grattafiori, Aaron and Dubey, Abhimanyu and Jauhri, Abhinav and others},
  year          = {2024},
  eprint        = {2407.21783},
  archivePrefix = {arXiv},
  primaryClass  = {cs.AI},
  url           = {https://arxiv.org/abs/2407.21783}
}

@inproceedings{lewis2021retrievalaugmentedgenerationknowledgeintensivenlp,
 author = {Lewis, Patrick and Perez, Ethan and Piktus, Aleksandra and Petroni, Fabio and Karpukhin, Vladimir and Goyal, Naman and K\"{u}ttler, Heinrich and Lewis, Mike and Yih, Wen-tau and Rockt\"{a}schel, Tim and Riedel, Sebastian and Kiela, Douwe},
 booktitle = {Advances in Neural Information Processing Systems},
 pages = {9459--9474},
 title = {Retrieval-Augmented Generation for Knowledge-Intensive NLP Tasks},
 volume = {33},
 year = {2020}
}

@inproceedings{Guu2020RAG,
author = {Guu, Kelvin and Lee, Kenton and Tung, Zora and Pasupat, Panupong and Chang, Ming-Wei},
title = {REALM: retrieval-augmented language model pre-training},
year = {2020},
booktitle = {International Conference on Machine Learning},
articleno = {368},
numpages = {10},
}

@inproceedings{Khandelwal2020knnlm,
  author       = {Urvashi Khandelwal and
                  Omer Levy and
                  Dan Jurafsky and
                  Luke Zettlemoyer and
                  Mike Lewis},
  title        = {Generalization through Memorization: Nearest Neighbor Language Models},
  booktitle    = {International Conference on Learning Representations (ICLR)},
  year         = {2020},
}

@article{ram-etal-2023-context,
    title = "In-Context Retrieval-Augmented Language Models",
    author = "Ram, Ori  and
      Levine, Yoav  and
      Dalmedigos, Itay  and
      Muhlgay, Dor  and
      Shashua, Amnon  and
      Leyton-Brown, Kevin  and
      Shoham, Yoav",
    journal = "Transactions of the Association for Computational Linguistics",
    year = "2023",
    pages = "1316--1331",
}

@inproceedings{selfrag,
  author       = {Akari Asai and
                  Zeqiu Wu and
                  Yizhong Wang and
                  Avirup Sil and
                  Hannaneh Hajishirzi},
  title        = {Self-RAG: Learning to Retrieve, Generate, and Critique through Self-Reflection},
  booktitle    = {The Twelfth International Conference on Learning Representations,
                  {ICLR} 2024, Vienna, Austria, May 7-11, 2024},
  publisher    = {OpenReview.net},
  year         = {2024},
  url          = {https://openreview.net/forum?id=hSyW5go0v8},
}

@misc{drozdov2023paradepassagerankingusing,
      title={PaRaDe: Passage Ranking using Demonstrations with Large Language Models}, 
      author={Andrew Drozdov and Honglei Zhuang and Zhuyun Dai and Zhen Qin and Razieh Rahimi and Xuanhui Wang and Dana Alon and Mohit Iyyer and Andrew McCallum and Donald Metzler and Kai Hui},
      year={2023},
      eprint={2310.14408},
      archivePrefix={arXiv},
      primaryClass={cs.IR},
      url={https://arxiv.org/abs/2310.14408}, 
}

@inproceedings{sachan2023improvingpassageretrievalzeroshot,
  author       = {Devendra Singh Sachan and
                  Mike Lewis and
                  Mandar Joshi and
                  Armen Aghajanyan and
                  Wen{-}tau Yih and
                  Joelle Pineau and
                  Luke Zettlemoyer},
  editor       = {Yoav Goldberg and
                  Zornitsa Kozareva and
                  Yue Zhang},
  title        = {Improving Passage Retrieval with Zero-Shot Question Generation},
  booktitle    = {Proceedings of the 2022 Conference on Empirical Methods in Natural
                  Language Processing, {EMNLP} 2022, Abu Dhabi, United Arab Emirates,
                  December 7-11, 2022},
  pages        = {3781--3797},
  publisher    = {Association for Computational Linguistics},
  year         = {2022},
  url          = {https://doi.org/10.18653/v1/2022.emnlp-main.249},
  doi          = {10.18653/V1/2022.EMNLP-MAIN.249},
}

@inproceedings{qin-etal-2024-large,
    title = "Large Language Models are Effective Text Rankers with Pairwise Ranking Prompting",
    author = "Qin, Zhen  and
      Jagerman, Rolf  and
      Hui, Kai  and
      Zhuang, Honglei  and
      Wu, Junru  and
      Yan, Le  and
      Shen, Jiaming  and
      Liu, Tianqi  and
      Liu, Jialu  and
      Metzler, Donald  and
      Wang, Xuanhui  and
      Bendersky, Michael",
    editor = "Duh, Kevin  and
      Gomez, Helena  and
      Bethard, Steven",
    booktitle = "Findings of the Association for Computational Linguistics: NAACL 2024",
    month = jun,
    year = "2024",
    address = "Mexico City, Mexico",
    publisher = "Association for Computational Linguistics",
    url = "https://aclanthology.org/2024.findings-naacl.97/",
    doi = "10.18653/v1/2024.findings-naacl.97",
    pages = "1504--1518",
    
}

@inproceedings{sun-etal-2023-chatgpt,
    title = "Is {C}hat{GPT} Good at Search? Investigating Large Language Models as Re-Ranking Agents",
    author = "Sun, Weiwei  and
      Yan, Lingyong  and
      Ma, Xinyu  and  others",
    editor = "Bouamor, Houda  and
      Pino, Juan  and
      Bali, Kalika",
    booktitle = "Proceedings of the 2023 Conference on Empirical Methods in Natural Language Processing",
    month = dec,
    year = "2023",
    address = "Singapore",
    publisher = "Association for Computational Linguistics",
    url = "https://aclanthology.org/2023.emnlp-main.923/",
    doi = "10.18653/v1/2023.emnlp-main.923",
    pages = "14918--14937",
}

@misc{ma2023zeroshotlistwisedocumentreranking,
      title={Zero-Shot Listwise Document Reranking with a Large Language Model}, 
      author={Xueguang Ma and Xinyu Zhang and Ronak Pradeep and Jimmy Lin},
      year={2023},
      eprint={2305.02156},
      archivePrefix={arXiv},
      primaryClass={cs.IR},
      url={https://arxiv.org/abs/2305.02156}, 
}

@inproceedings{Cao2007LTR,
  author       = {Zhe Cao and
                  Tao Qin and
                  Tie{-}Yan Liu and
                  Ming{-}Feng Tsai and
                  Hang Li},
  editor       = {Zoubin Ghahramani},
  title        = {Learning to rank: from pairwise approach to listwise approach},
  booktitle    = {Machine Learning, Proceedings of the Twenty-Fourth International Conference
                  {(ICML} 2007), Corvallis, Oregon, USA, June 20-24, 2007},
  series       = {{ACM} International Conference Proceeding Series},
  volume       = {227},
  pages        = {129--136},
  publisher    = {{ACM}},
  year         = {2007},
  url          = {https://doi.org/10.1145/1273496.1273513},
  doi          = {10.1145/1273496.1273513},
}

@article{nogueira2019passage,
  author       = {Rodrigo Nogueira and
                  Kyunghyun Cho},
  title        = {Passage Re-ranking with {BERT}},
  journal      = {CoRR},
  volume       = {abs/1901.04085},
  year         = {2019},
  url          = {http://arxiv.org/abs/1901.04085},
  eprinttype    = {arXiv},
  eprint       = {1901.04085},
}

@article{search-r1,
  author       = {Bowen Jin and
                  Hansi Zeng and
                  Zhenrui Yue and
                  Dong Wang and
                  Hamed Zamani and
                  Jiawei Han},
  title        = {Search-R1: Training LLMs to Reason and Leverage Search Engines with
                  Reinforcement Learning},
  journal      = {CoRR},
  volume       = {abs/2503.09516},
  year         = {2025},
  url          = {https://doi.org/10.48550/arXiv.2503.09516},
  doi          = {10.48550/ARXIV.2503.09516},
  eprinttype    = {arXiv},
  eprint       = {2503.09516},
}

@inproceedings{dpr,
  author       = {Vladimir Karpukhin and
                  Barlas Oguz and
                  Sewon Min and
                  Patrick Lewis and
                  Ledell Wu and
                  Sergey Edunov and
                  Danqi Chen and
                  Wen{-}tau Yih},
  editor       = {Bonnie Webber and
                  Trevor Cohn and
                  Yulan He and
                  Yang Liu},
  title        = {Dense Passage Retrieval for Open-Domain Question Answering},
  booktitle    = {Proceedings of the 2020 Conference on Empirical Methods in Natural
                  Language Processing, {EMNLP} 2020, Online, November 16-20, 2020},
  pages        = {6769--6781},
  publisher    = {Association for Computational Linguistics},
  year         = {2020},
  url          = {https://doi.org/10.18653/v1/2020.emnlp-main.550},
  doi          = {10.18653/V1/2020.EMNLP-MAIN.550},
}

@misc{dacl-rag,
      title={DACL-RAG: Data Augmentation Strategy with Curriculum Learning for Retrieval-Augmented Generation}, 
      author={Shaohan Wang and Licheng Zhang and Zheren Fu and Zhendong Mao and Yongdong Zhang},
      year={2025},
      eprint={2505.10493},
      archivePrefix={arXiv},
      primaryClass={cs.CL},
      url={https://arxiv.org/abs/2505.10493}, 
}

@article{search-o1,
  author       = {Xiaoxi Li and
                  Guanting Dong and
                  Jiajie Jin and
                  Yuyao Zhang and
                  Yujia Zhou and
                  Yutao Zhu and
                  Peitian Zhang and
                  Zhicheng Dou},
  title        = {Search-o1: Agentic Search-Enhanced Large Reasoning Models},
  journal      = {CoRR},
  volume       = {abs/2501.05366},
  year         = {2025},
  url          = {https://doi.org/10.48550/arXiv.2501.05366},
  doi          = {10.48550/ARXIV.2501.05366},
  eprinttype    = {arXiv},
  eprint       = {2501.05366},
}

@inproceedings{IRCoT,
  author       = {Harsh Trivedi and
                  Niranjan Balasubramanian and
                  Tushar Khot and
                  Ashish Sabharwal},
  editor       = {Anna Rogers and
                  Jordan L. Boyd{-}Graber and
                  Naoaki Okazaki},
  title        = {Interleaving Retrieval with Chain-of-Thought Reasoning for Knowledge-Intensive
                  Multi-Step Questions},
  booktitle    = {Proceedings of the 61st Annual Meeting of the Association for Computational
                  Linguistics (Volume 1: Long Papers), {ACL} 2023, Toronto, Canada,
                  July 9-14, 2023},
  pages        = {10014--10037},
  publisher    = {Association for Computational Linguistics},
  year         = {2023},
  url          = {https://doi.org/10.18653/v1/2023.acl-long.557},
  doi          = {10.18653/V1/2023.ACL-LONG.557},
}

@misc{cao2025singlepassdocumentscanningquestion, title={Single-Pass Document Scanning for Question Answering}, author={Weili Cao and Jianyou Wang and Youze Zheng and Longtian Bao and Qirui Zheng and Taylor Berg-Kirkpatrick and Ramamohan Paturi and Leon Bergen}, year={2025}, eprint={2504.03101}, archivePrefix={arXiv}, primaryClass={cs.CL}, url={https://arxiv.org/abs/2504.03101}, }
\bibliographystyle{icml2026}

%%%%%%%%%%%%%%%%%%%%%%%%%%%%%%%%%%%%%%%%%%%%%%%%%%%%%%%%%%%%%%%%%%%%%%%%%%%%%%%
%%%%%%%%%%%%%%%%%%%%%%%%%%%%%%%%%%%%%%%%%%%%%%%%%%%%%%%%%%%%%%%%%%%%%%%%%%%%%%%
% APPENDIX
%%%%%%%%%%%%%%%%%%%%%%%%%%%%%%%%%%%%%%%%%%%%%%%%%%%%%%%%%%%%%%%%%%%%%%%%%%%%%%%
%%%%%%%%%%%%%%%%%%%%%%%%%%%%%%%%%%%%%%%%%%%%%%%%%%%%%%%%%%%%%%%%%%%%%%%%%%%%%%%
\newpage
\appendix
\onecolumn

\section{Appendix}

\subsection{Algorithm}

We present our algorithm of \texttt{BAR-RAG} in Algorithm \ref{alg:BAR-RAG}.

\begin{algorithm*}[htbp]
\caption{Boundary-aware Evidence Selection Training (\texttt{BAR-RAG})}
\label{alg:BAR-RAG}
\begin{algorithmic}[1]
\REQUIRE Training dataset $\mathcal{D}$, retriever $\mathcal{R}$, selector $\pi_r$, generator $\pi_g$
\REQUIRE Target correctness $c$, reward threshold $\delta$, rollouts $K$, evidence samples $M$
\REQUIRE Training iterations $T$

\STATE \textbf{// Training-time filtering}
\STATE $\mathcal{D}_{\text{filt}} \leftarrow \textsc{Filter}(\mathcal{D})$

\FOR{$t = 1$ to $T$}
    \STATE \textbf{// Stage 1: Selector Training (generator frozen)}
    \FOR{each query $q$ in batch from $\mathcal{D}_{\text{filt}}$}
        \STATE $C \leftarrow \mathcal{R}(q)$
        \STATE Sample $M$ evidence sets $\{S^{(m)}\}_{m=1}^{M}$ from $\pi_r$
        \FOR{each $S^{(m)}$}
            \STATE Sample $K$ answers from $\pi_g(\cdot \mid q, S^{(m)})$
            \STATE Estimate $\hat{p}(S^{(m)})$ and compute selector reward $R_r(S^{(m)})$
        \ENDFOR
        \STATE Update selector $\pi_r$ via GRPO
    \ENDFOR

    \STATE \textbf{// Stage 2: Generator Training (selector frozen)}
    \FOR{each query $q$ in batch from $\mathcal{D}$}
        \STATE $C \leftarrow \mathcal{R}(q)$
        \STATE $S \leftarrow \pi_r(q, C)$
        \STATE Sample $K$ answers from $\pi_g(\cdot \mid q, S)$
        \STATE Update generator $\pi_g$ via GRPO
    \ENDFOR
\ENDFOR
\end{algorithmic}
\end{algorithm*}

\subsection{Comparison with Recent Reasoning and Agentic RAG Methods}
\label{app:agentic_comparison}

To further contextualize \texttt{BAR-RAG} within the rapidly evolving landscape of reasoning-centric and agentic RAG systems, we compare against several recent representative methods, including Search-o1 \cite{search-o1}, Search-R1 \cite{search-r1}, and DynamicRAG \cite{dynamicrag}. \footnote{As these methods vary widely in backbone models, inference-time computation, and retrieval interfaces, a strictly controlled apples-to-apples comparison is not always possible.
Therefore, this table is intended to provide a qualitative and contextual comparison rather than a claim of direct superiority.}
These approaches emphasize explicit multi-step search, iterative retrieval, or reinforcement-learning-based reranking at inference time, and are often evaluated under substantially different retrieval and search budgets.

In particular, we group results by backbone scale following the structure of Table \ref{tab:agentic_comparison}, and report the best-performing iteration of \texttt{BAR-RAG} for each backbone.
Importantly, \texttt{BAR-RAG} incurs no additional inference-time overhead beyond standard RAG, whereas the compared reasoning and agentic methods typically rely on multi-round search or tool invocation during inference.

\begin{table*}[t]
\centering
\caption{Comparison with recent reasoning and agentic RAG methods (EM, \%).
Results are grouped by backbone following Table~1, and BAR-RAG reports the best iteration for each backbone.
Search-o1 and Search-R1 results are taken directly from the original papers.
DynamicRAG results are obtained by re-running the authors' released checkpoint under our evaluation pipeline, as its original setting is not fully aligned with ours.
Notably, Search-R1 employs substantially different retrieval configurations on multi-hop QA (HotpotQA, 2WikiMultiHopQA, MuSiQue and Bamboogle), involving multiple rounds of search with multiple documents per round, whereas our setting uses a single-shot top-$k$ retrieval (top-5) without iterative search.
As a result, Search-R1 results on these datasets (marked with $^{\dagger}$) are not directly comparable to \texttt{BAR-RAG} and should be interpreted with caution.
}
\label{tab:agentic_comparison}
\resizebox{0.95\textwidth}{!}{
\begin{tabular}{lcccccccc}
\toprule
\textbf{Method} & \textbf{NQ} & \textbf{TriviaQA} & \textbf{PopQA} & \textbf{HotpotQA} & \textbf{2Wiki} & \textbf{MuSiQue} & \textbf{Bamboogle} & \textbf{Avg.} \\
\midrule

\multicolumn{9}{c}{\cellcolor{gray!20}\textbf{Qwen2.5-3B-Instruct}} \\
\midrule
Search-o1              & 23.8 & 47.2 & 26.2 & 22.1 & 21.8 & 5.4  & 32.0 & 25.5 \\
Search-R1     & 34.1 & 54.5 & 37.8 & 32.4$\dagger$ & \textbf{31.9}$\dagger$ & \textbf{10.3}$\dagger$ & \textbf{26.4}$\dagger$ & 32.5 \\
\textbf{\texttt{BAR-RAG}}  & \textbf{41.5} & \textbf{61.3} & \textbf{43.4} & \textbf{33.2} & 26.4 & 7.4 & 26.3& \textbf{34.3} \\
\midrule

\multicolumn{9}{c}{\cellcolor{gray!20}\textbf{Qwen2.5-7B-Instruct}} \\
\midrule
Search-o1              & 15.1 & 44.3 & 13.1 & 18.7 & 17.6 & 5.8  & 29.6 & 20.6 \\
Search-R1     & 39.3 & 61.0 & 39.7 & 37.0$\dagger$ & \textbf{41.4}$\dagger$ & \textbf{14.6}$\dagger$ & 36.8$\dagger$ & 38.5 \\
\textbf{\texttt{BAR-RAG}}   & \textbf{46.9} & \textbf{64.5} & \textbf{46.9} & \textbf{38.8} & 29.8 & 9.1 & \textbf{39.6} & \textbf{39.1} \\
\midrule

\multicolumn{9}{c}{\cellcolor{gray!20}\textbf{LLaMA-3.1-8B-Instruct}} \\
\midrule
DynamicRAG             & 46.4 & 57.5 & 36.7 & 34.2 & --   & --   & --   & \\
\textbf{\texttt{BAR-RAG}}  & \textbf{49.5} & \textbf{67.3} & \textbf{48.6} & \textbf{41.2} & 33.0 & 12.0 & 33.3 & 40.7 \\
\bottomrule
\end{tabular}
}
\end{table*}

\subsection{Counterfactual Evidence Dependence Analysis}
\label{app:counterfactual}

A potential concern with generator-aware evidence selection is whether performance gains
arise from genuine evidence use or from self-confirmation effects induced by generator
feedback.
To directly test whether the model’s predictions causally depend on the evidence it cites,
we perform a counterfactual evidence-dependence analysis.

For each test example, we first run the model under the standard setting (\textsc{Full})
using the top-$k$ retrieved documents, and record the set of documents cited in the model’s
generation.
We then construct two counterfactual variants while keeping the question and decoding
settings fixed:
(i) \textsc{Remove-Cited}, where all documents cited by the model in the original generation
are removed from the input, and
(ii) \textsc{Keep-Only-Cited}, where only the cited documents are retained and all other
retrieved documents are discarded.
The cited document set is always determined from the original \textsc{Full} generation and
kept fixed across counterfactual runs.

Table~\ref{tab:counterfactual_cited} reports Exact Match (EM) results on NQ, HotpotQA, and
Bamboogle using Qwen2.5-7B-Instruct.
Across all datasets, removing the cited evidence causes a large performance drop, with
$\Delta_{\mathrm{rm}}$ exceeding 24 EM on all three benchmarks.
This sharp degradation indicates that correct predictions critically rely on the documents
explicitly referenced by the model.

In contrast, retaining only the cited documents largely preserves performance and in some
cases slightly improves it (e.g., NQ and Bamboogle), suggesting that the cited documents
form a compact and sufficient support set, while additional retrieved documents often act
as noise.

\begin{table}[b]
\centering
\caption{Counterfactual evidence-dependence analysis on Qwen2.5-7B-Instruct.
\textsc{Remove-Cited} removes all documents cited by the model in the original generation,
while \textsc{Keep-Only-Cited} retains only the cited documents.
\(\Delta_{\mathrm{rm}}\) denotes the EM drop from \textsc{Full} to \textsc{Remove-Cited}.}
\label{tab:counterfactual_cited}
\begin{tabular}{lcccc}
\toprule
\textbf{Dataset} &
\textbf{Full} &
\textbf{Remove-Cited} &
\textbf{Keep-Only-Cited} &
\(\boldsymbol{\Delta_{\mathrm{rm}}}\) \\
\midrule
NQ         & 46.9 & 20.4 & 47.3 & 26.5 \\
HotpotQA  & 38.8 & 12.2 & 38.5 & 26.6 \\
Bamboogle & 39.6 & 15.3 & 40.1 & 24.3 \\
\bottomrule
\end{tabular}
\end{table}

\subsection{Extended Related Work}
\label{sec:extended_rw}

\subsubsection{Retrieval-Augmented Generation}
Retrieval-Augmented Generation (RAG) has emerged as a general framework for grounding language model outputs in external knowledge sourceAR
s, mitigating hallucination and enabling dynamic knowledge updates.
Early formulations combine neural retrieval with sequence-to-sequence generation for knowledge-intensive tasks \cite{lewis2021retrievalaugmentedgenerationknowledgeintensivenlp, Guu2020RAG, dacl-rag, cao2025singlepassdocumentscanningquestion}.
Beyond document-level retrieval, several works explore datastore-based or token-level augmentation.
For example, kNN-LM \cite{Khandelwal2020knnlm} augments next-token prediction with nearest-neighbor retrieval over a large datastore, while subsequent extensions improve efficiency and contextualization \cite{ram-etal-2023-context}.

Recent research has focused on tighter integration between retrieval and generation.
Some approaches train retrieval and generation components end-to-end, while others introduce explicit control mechanisms to decide when retrieval should occur.
Self-RAG \cite{selfrag} equips the generator with a learned critic that reflects on intermediate generations and dynamically triggers retrieval.
These methods primarily focus on retrieval timing or integration strategy, whereas our work complements them by addressing the orthogonal problem of evidence selection quality under fixed retrieval budgets.

\subsubsection{Document Reranking}
Reranking is a long-standing component in information retrieval pipelines, aiming to refine initial retrieval results before downstream consumption.
Traditional learning-to-rank approaches optimize pairwise or listwise objectives over hand-crafted features \cite{Cao2007LTR}.
With the advent of pretrained language models, neural rerankers based on cross-encoder architectures have become the dominant paradigm, jointly encoding query--document pairs to compute relevance scores with fine-grained token interactions \cite{nogueira2019passage}.

Large language models have recently been explored as rerankers, leveraging their reasoning capabilities to assess document usefulness.
Prior work spans multiple granularities, including pointwise relevance prediction or likelihood estimation \cite{drozdov2023paradepassagerankingusing, sachan2023improvingpassageretrievalzeroshot}, pairwise comparison of candidate documents \cite{qin-etal-2024-large}, and listwise ranking that directly outputs document permutations \cite{sun-etal-2023-chatgpt, ma2023zeroshotlistwisedocumentreranking}.
Many of these approaches operate in a zero-shot or weakly supervised setting and optimize relevance-based objectives.

Several recent works incorporate reinforcement learning to optimize reranking policies.
DynamicRAG \cite{dynamicrag} formulates reranking as a sequential decision process and uses generator feedback as a reward signal to dynamically determine both the ordering and the number of documents.
While effective, such approaches typically rely on relevance or answer-quality-based rewards.
In contrast, our method introduces an explicit competence-aware objective that directly models evidence difficulty, targeting hard-but-solvable evidence sets that maximize learning signal for downstream generator training.

\subsection{Training Implementation Details}
\label{sec:impl_details}

We use \textit{Qwen2.5-3B-Instruct}, \textit{Qwen2.5-7B-Instruct}, and \textit{LLaMA-3.1-8B-Instruct} as backbone models for both the selector and the generator.
In both training stages, we adopt parameter-efficient fine-tuning using LoRA adapters with rank 32 and scaling factor $\alpha=16$, while keeping all backbone parameters frozen.

For retrieval, we use \textit{E5} as the dense retriever to obtain a fixed pool of top-$n=25$ candidate documents for each query.
During selector training, we sample $M=8$ candidate evidence sets per query, each consisting of $k=5$ documents.

Each sampled evidence set is evaluated using $K=10$ generator rollouts to estimate the correctness probability $\hat{p}(S)$.
A rollout is deemed correct if its final generator reward exceeds a threshold $\delta=0.8$.
The selector reward targets a correctness level of $c=0.5$, with boundary and relevance weights $\lambda_{\mathrm{bdy}}=1.0$ and $\lambda_{\mathrm{rel}}=0.2$, respectively.
We use a relevance temperature $\tau=10.0$ and apply a count penalty of $\alpha=0.5$ per document deviation from the target size $k^*=5$, capped at $P_{\max}=1.0$.

The generator reward combines answer accuracy and citation quality with weights $\lambda_{\mathrm{acc}}=0.8$ and $\lambda_{\mathrm{cite}}=0.2$.
The accuracy reward weights token-level F1 and exact match as $\beta_1=0.7$ and $\beta_2=0.3$, respectively.
The citation reward targets $n^*=2$ cited documents.

Both stages are optimized using GRPO with a learning rate of $4\times10^{-6}$, cosine learning rate decay with 2\% warmup, clipping parameter $\epsilon=0.2$, batch size 8, and KL regularization coefficient $0.001$.
Training is conducted for three iterations on $8\times$A100 (40GB) GPUs using bfloat16 mixed-precision training with gradient checkpointing.

\subsection{Training-time Filtering Details}
\label{app:filtering}

We now provide implementation details for the filtering procedure described in Section~\ref{sec:filter}.
For each training query, we sample $N$ reranker rollouts and $K$ generator rollouts per evidence set.
Generator outputs are judged correct if their total reward exceeds a fixed threshold $\delta$, consistent with the correctness definition used during selector training.

In practice, we compute the mean and variance of empirical correctness across reranker rollouts.
Queries with near-zero variance or near-deterministic outcomes are removed, as they correspond to trivially easy or unanswerable cases.
Unless otherwise stated, we retain queries whose mean correctness satisfies
$\mu_q \in [m_{\min}, m_{\max}]$ and whose variance exceeds $v_{\min}$.

In all experiments, we use $N = 8$ reranker rollouts and $K = 10$ generator rollouts.
We set the correctness threshold to $\delta = 0.5$.
The mean correctness bounds are $m_{\min} = 0.25$ and $m_{\max} = 0.85$, and the minimum variance threshold is $v_{\min} = 0.02$.
This filtering step is applied only to selector training and does not affect generator training.

\subsection{Training Data}
\label{appendix:training_data}

Following \citet{search-r1}, we construct the training set by merging the training splits of Natural Questions (NQ)~\citep{nq} and HotpotQA~\citep{hotpotqa}. NQ provides single-hop factoid questions derived from real Google search queries, while HotpotQA contributes multi-hop questions that require reasoning over multiple Wikipedia passages. This combination ensures coverage of both simple retrieval scenarios and complex multi-step reasoning tasks.

For the retrieval corpus, we use the 2018 Wikipedia dump~\citep{dpr}, which contains approximately 21 million passages. We employ E5~\citep{e5} as the dense retriever. For each query, we retrieve the top-25 passages to form a fixed retrieval pool, from which the selector learns to compose evidence subsets during training.

Table~\ref{tab:data_statistics} summarizes the dataset statistics. The combined training set contains 87,925 question-answer pairs. Each training instance consists of a question $q$, the ground-truth answer $a$, and a retrieval pool $\mathcal{D}_q$ containing the top-25 passages retrieved for $q$. During \texttt{BAR-RAG} training, the selector operates over this fixed retrieval pool, learning to compose evidence subsets that challenge the generator near its competence boundary.

We evaluate on seven benchmark datasets to assess both in-domain and out-of-domain generalization: (1) \textbf{In-domain}: NQ and HotpotQA; (2) \textbf{Out-of-domain}: TriviaQA~\citep{triviaqa}, PopQA~\citep{popqa}, 2WikiMultiHopQA~\citep{2wikimqa}, Musique~\citep{musique}, and Bamboogle~\citep{bamboogle}. Following \citet{search}, we use Exact Match (EM) as the evaluation metric.

\begin{table}[t]
\centering
\caption{Dataset statistics for training and evaluation.}
\label{tab:data_statistics}
\resizebox{0.8\columnwidth}{!}{%
\begin{tabular}{llrrr}
\toprule
\textbf{Split} & \textbf{Dataset} & \textbf{\#Examples} & \textbf{Task Type} & \textbf{Domain} \\
\midrule
\multirow{3}{*}{Train} & NQ & 79,168 & Single-hop & In-domain \\
& HotpotQA & 8,757 & Multi-hop & In-domain \\
& \textit{Total} & \textit{87,925} & -- & -- \\
\midrule
\multirow{7}{*}{Eval} & NQ & 3,610 & Single-hop & In-domain \\
& HotpotQA & 7,405 & Multi-hop & In-domain \\
\cmidrule{2-5}
& TriviaQA & 11,313 & Single-hop & Out-of-domain \\
& PopQA & 14,267 & Single-hop & Out-of-domain \\
& 2WikiMultiHopQA & 12,576 & Multi-hop & Out-of-domain \\
& Musique & 2,417 & Multi-hop & Out-of-domain \\
& Bamboogle & 125 & Multi-hop & Out-of-domain \\
\bottomrule
\end{tabular}
}
\end{table}

\subsection{Case Study}

Table~\ref{tab:case_study_BAR-RAG} presents a representative case study illustrating how \texttt{BAR-RAG} progressively hardens evidence composition across training iterations while operating over a fixed top-25 retrieval pool.

In Iteration~1, the selector surfaces the two \textcolor{blue}{[GOLDEN]} documents at the top of the evidence list. One document explicitly identifies Claudio L\'{o}pez as a retired Argentine forward who played as a main attacking player for Valencia CF, while the other confirms his role as a regular starter in Valencia’s attacking line during the relevant period. The remaining documents are largely \textcolor{red}{[IRRELEVANT]} and do not introduce strong competing signals. As a result, the generator can answer the question correctly with minimal reasoning effort, relying primarily on direct evidence aggregation.

In Iteration~2, the same two golden documents remain present but are no longer adjacent. They are interleaved with \textcolor{orange}{[MISLEADING]} documents that are highly relevant at the surface level, such as profiles of Mario Kempes or summaries of Argentine forwards in La Liga. These distractors satisfy several query attributes (e.g., nationality, position, club association) but fail to jointly satisfy all constraints. Consequently, positional heuristics or shallow relevance matching become unreliable, and correct prediction requires identifying and combining the truly decisive evidence.

By Iteration~3, \texttt{BAR-RAG} further increases structural difficulty by pushing the two golden documents deeper into the evidence set and surrounding them with multiple misleading but plausible alternatives. The shortcut evidence remains visible but becomes insufficient: although figures such as Mario Kempes match many individual query facets, they do not simultaneously satisfy the conjunction of being retired, Argentine, a forward, and a main player for Valencia CF during the specified era. Only by consistently tracking entity identity across the dispersed golden documents can the generator arrive at the correct answer.

Across iterations, task solvability is preserved, as the same two golden documents are always present. However, the evidence structure is progressively hardened: decisive information is no longer top-ranked or contiguous, and misleading cues increasingly dominate surface relevance. This case study demonstrates that \texttt{BAR-RAG} improves robustness not by introducing new evidence, but by reshaping the composition and ordering of existing retrieval results to suppress shortcut reasoning and induce genuine multi-hop integration near the generator’s competence boundary.

\begin{table*}[t]
\centering
\caption{Case study illustrating boundary-aware evidence selection from a fixed top-25 retrieval pool.
Document IDs correspond to original retrieval ranks.
Across iterations, the same two \textcolor{blue}{[GOLDEN]} documents remain necessary to answer the question, while their relative positions are progressively dispersed and surrounded by \textcolor{orange}{[MISLEADING]} but topically relevant noise and \textcolor{red}{[IRRELEVANT]} documents.
This structured hardening preserves solvability while forcing multi-hop reasoning near the generator's competence boundary.}
\resizebox{\textwidth}{!}{%
\begin{tabular}{p{0.32\textwidth}p{0.32\textwidth}p{0.32\textwidth}}
\toprule
\multicolumn{3}{l}{\textbf{Question:} Which retired Argentine footballer who played as a forward was a main player for Valencia CF?} \\
\midrule
\textbf{Iteration 1} 
& \textbf{Iteration 2} 
& \textbf{Iteration 3} \\
\midrule

% ---------------- Iteration 1 ----------------
\begin{minipage}[t]{\linewidth}
\textbf{(Doc 1)} \textcolor{blue}{[GOLDEN]} \newline
\textbf{Title:} Claudio L\'{o}pez — Career Summary \newline
\textbf{Content:} Claudio L\'{o}pez is a retired Argentine footballer who played as a forward and was a main attacking player for Valencia CF. \vspace{6pt}

\textbf{(Doc 2)} \textcolor{blue}{[GOLDEN]} \newline
\textbf{Title:} Valencia CF Squad (1998--2000) \newline
\textbf{Content:} Valencia relied on Claudio L\'{o}pez as a regular starter in their attacking line during this period. \vspace{6pt}

\textbf{(Doc 3)} \textcolor{red}{[IRRELEVANT]} \newline
\textbf{Title:} Valencia CF History (1990s) \vspace{6pt}

\textbf{(Doc 4)} \textcolor{red}{[IRRELEVANT]} \newline
\textbf{Title:} Notable Footballers in La Liga \vspace{6pt}

\textbf{(Doc 5)} \textcolor{orange}{[MISLEADING]}  \newline
\textbf{Title:} Overview of Spanish Football Clubs
\end{minipage}

&
% ---------------- Iteration 2 ----------------
\begin{minipage}[t]{\linewidth}
\textbf{(Doc 1)} \textcolor{blue}{[GOLDEN]} \newline
\textbf{Title:} Claudio L\'{o}pez — Career Summary \newline
\textbf{Content:} Claudio L\'{o}pez played as a main forward for Valencia CF in the late 1990s. \vspace{6pt}

\textbf{(Doc 5)} \textcolor{orange}{[MISLEADING]} \newline
\textbf{Title:} Mario Kempes — Career Overview \newline
\textbf{Content:} Mario Kempes is a retired Argentine forward and one of Valencia CF's most iconic historical players. \vspace{6pt}

\textbf{(Doc 2)} \textcolor{blue}{[GOLDEN]} \newline
\textbf{Title:} Valencia CF Squad (1998--2000) \newline
\textbf{Content:} Valencia relied on a fast Argentine forward as a regular starter in their attacking system. \vspace{6pt}

\textbf{(Doc 7)} \textcolor{orange}{[MISLEADING]} \newline
\textbf{Title:} Argentine Forwards in La Liga \newline
\textbf{Content:} Several Argentine forwards, including Kempes and others, played important roles in La Liga clubs. \vspace{6pt}

\textbf{(Doc 8)} \textcolor{red}{[IRRELEVANT]} \newline
\textbf{Title:} History of La Liga Stadiums
\end{minipage}

&
% ---------------- Iteration 3 ----------------
\begin{minipage}[t]{\linewidth}
\textbf{(Doc 5)} \textcolor{orange}{[MISLEADING]} \newline
\textbf{Title:} Mario Kempes — Career Overview \newline
\textbf{Content:} Kempes is remembered as one of the most influential Argentine forwards in Spanish football. \vspace{6pt}

\textbf{(Doc 11)} \textcolor{orange}{[MISLEADING]} \newline
\textbf{Title:} Argentine Legends in La Liga \newline
\textbf{Content:} Several Argentine forwards achieved legendary status at Spanish clubs. \vspace{6pt}

\textbf{(Doc 1)} \textcolor{blue}{[GOLDEN]} \newline
\textbf{Title:} Claudio L\'{o}pez — Career Summary \newline
\textbf{Content:} Claudio L\'{o}pez is a retired Argentine forward who played for Valencia CF. \vspace{6pt}

\textbf{(Doc 14)} \textcolor{orange}{[MISLEADING]} \newline
\textbf{Title:} Valencia CF Legends \newline
\textbf{Content:} Valencia CF has featured many historically significant attacking players. \vspace{6pt}

\textbf{(Doc 2)} \textcolor{blue}{[GOLDEN]} \newline
\textbf{Title:} Valencia CF Squad (1998--2000) \newline
\textbf{Content:} Valencia’s main attacking options during this era included a fast Argentine forward.
\end{minipage}

\\
\midrule
\multicolumn{3}{l}{\textbf{Answer:} Claudio Javier L\'{o}pez} \\
\bottomrule
\end{tabular}
}
\label{tab:case_study_BAR-RAG}
\end{table*}

\newpage

\subsection{Prompt}

This section presents the prompt templates we used for the generator and selector, in Table \ref{tab:generator} and Table \ref{tab:selector}, respectively.

\begin{table}[htbp]
\centering
\caption{Prompt template for reasoning generator.}
\begin{minipage}{0.9\columnwidth}
    \centering
    \begin{tcolorbox}[title=Reasoning Generator Prompt]
        You are given a question and a set of retrieved documents. Your task is to answer the question \textbf{using only information from the retrieved documents}. Even for yes/no questions, you must determine the answer by reasoning from factual evidence in the documents.

        \vspace{0.6em}
        \textbf{Output format (STRICT):}
        \begin{itemize}
            \item \texttt{<think>} A concise reasoning chain explaining how the answer is derived from the documents. Keep it brief (1--3 sentences). \texttt{</think>}
            \item \texttt{<answer>} The final answer. \texttt{</answer>}
        \end{itemize}

        \vspace{0.6em}
        \textbf{Evidence citation rule:}
        \begin{itemize}
            \item Whenever you use a piece of evidence from the documents in your reasoning, you \textbf{must} cite it inline as \texttt{Doc [i]}.
            \item You may cite one or multiple documents, but only cite documents that are actually relevant.
        \end{itemize}

        \vspace{0.6em}
        \textbf{Answer rules:}
        \begin{itemize}
            \item The answer should be a \textbf{short phrase} directly supported by the retrieved documents.
            \item Do \textbf{not} introduce external knowledge or assumptions.
            \item Do \textbf{not} output anything outside \texttt{<think>} and \texttt{<answer>}.
        \end{itemize}

        \vspace{0.6em}
        \textbf{Example (follow the style only):}\\
        \texttt{<think> Doc [1] states that Future Ted serves as the show's narrator, and Doc [4] confirms that the narrator is voiced by Bob Saget. </think>}\\
        \texttt{<answer> Ted Mosby </answer>}

        \vspace{0.4em}
        \texttt{<Question>}\\
        \texttt{<Retrieved Documents>}
    \end{tcolorbox}
\end{minipage}
\label{tab:generator}
\end{table}

\begin{table}[htbp]
\centering
\caption{Prompt template for evidence selector.}
\begin{minipage}{0.9\columnwidth}
    \centering
    \begin{tcolorbox}[title=Evidence Selector Prompt]
        You are an expert evidence-set selector for RAG. Your goal is to select \textbf{exactly five} documents that make the question \textbf{answerable, but not trivial}. Prefer evidence sets that sit near the model's \textbf{competence boundary}: solvable with careful multi-step reasoning, yet not so direct that the answer is obvious from a single passage. You must follow the principles and output format strictly.

        \vspace{0.6em}
        \textbf{Principles:}
        \begin{enumerate}
            \item \textbf{Answerability (must-have):} The selected set must contain enough information to deduce the correct answer. Do \textbf{not} select sets that make the question impossible.
            \item \textbf{Non-triviality (must-have):} Avoid sets where one document directly states the answer with no integration needed. If a direct-answer passage is unavoidable for solvability, include it \textbf{only together with} supporting/context passages that require cross-document integration.
            \item \textbf{Multi-hop integration:} Prefer sets that require combining at least \textbf{two} complementary clues (e.g., entity linking, temporal alignment, resolving aliases, chaining relations).
            \item \textbf{Controlled noise:} Mildly conflicting or distracting details are allowed if the set remains answerable; do not include documents that are irrelevant or make the set unsolvable.
            \item \textbf{Diversity:} Prefer complementary documents covering different parts of the reasoning chain, rather than near-duplicates.
        \end{enumerate}

        \vspace{0.6em}
        \textbf{Output format (STRICT):}
        \begin{itemize}
            \item \texttt{<think>} Briefly explain which documents contain key clues, how they complement each other, and why the set is answerable but requires integration. Keep it concise (2--4 sentences). \texttt{</think>}
            \item \texttt{<answer> [doc\_id1], [doc\_id2], [doc\_id3], [doc\_id4], [doc\_id5] </answer>}
        \end{itemize}

        \vspace{0.6em}
        \textbf{Rules:}
        \begin{itemize}
            \item Select \textbf{exactly 5} documents.
            \item In \texttt{<answer>}, list \textbf{only} the document identifiers in brackets, separated by commas.
            \item Do \textbf{not} output anything outside \texttt{<think>} and \texttt{<answer>}.
        \end{itemize}

        \vspace{0.6em}
        \textbf{Example (follow the style only):}\\
        \texttt{<think> Doc [3] provides the birthplace clue, Doc [7] gives a timeline, and Doc [12] resolves an alias; combining them is necessary. Doc [5] and Doc [9] add supporting context while introducing mild distraction, keeping the set solvable but non-trivial. </think>}\\
        \texttt{<answer> [3], [5], [7], [9], [12] </answer>}

        \vspace{0.4em}
        \texttt{<Question>}\\
        \texttt{<Candidate Documents (Top-$K$)>}
        
    \end{tcolorbox}
\end{minipage}
\label{tab:selector}
\end{table}

%%%%%%%%%%%%%%%%%%%%%%%%%%%%%%%%%%%%%%%%%%%%%%%%%%%%%%%%%%%%%%%%%%%%%%%%%%%%%%%
%%%%%%%%%%%%%%%%%%%%%%%%%%%%%%%%%%%%%%%%%%%%%%%%%%%%%%%%%%%%%%%%%%%%%%%%%%%%%%%

\end{document}